\documentclass[journal]{IEEEtran}
\usepackage{algorithmic}
\usepackage{textcomp}

\usepackage{url}
\usepackage{verbatim}
\usepackage{graphicx}
\def\BibTeX{{\rm B\kern-.05em{\sc i\kern-.025em b}\kern-.08em
    T\kern-.1667em\lower.7ex\hbox{E}\kern-.125emX}}
\usepackage{balance}

\usepackage{colortbl}  
\usepackage{array}   
\usepackage{cite}
\usepackage{mathrsfs}            
\usepackage{amsmath,amssymb,amsfonts}
\usepackage{multirow}

\usepackage{algorithm} 

\usepackage{enumerate}
\usepackage{enumitem} 
\usepackage{mathrsfs}            
\usepackage{multirow}
\usepackage{enumerate}
\usepackage{booktabs}
\usepackage{threeparttable}
\usepackage{subfigure}
\usepackage{tablefootnote}
\usepackage{listings}
\usepackage{hyperref}
\usepackage{bbding}
\usepackage{balance}
\usepackage{hhline} 

\usepackage{xpatch}
\usepackage{xcolor}
\usepackage{array}
\usepackage{verbatimbox}

\begin{document}

\title{Grounded by Experience: Generative Healthcare Prediction Augmented with Hierarchical Agentic Retrieval}

\author{Chuang~Zhao,
        Hui~Tang, Hongke Zhao, Xiaofang Zhou,~\IEEEmembership{Fellow, IEEE}, Xiaomeng~Li,~\IEEEmembership{Senior Member, IEEE}
        
\IEEEcompsocitemizethanks{
\IEEEcompsocthanksitem C. Zhao, H Tang, and X. Li are with the Department of Electronic and Computer Engineering, The Hong Kong University of Science and Technology, Hong Kong, SAR, China;  (e-mail: czhaobo@connect.ust.hk, eehtang@ust.hk, eexmli@ust.hk).  X. Li is the corresponding author.
\vspace{-0.2em}
\IEEEcompsocthanksitem H. Zhao is with the College of Management and Economics, Laboratory of Computation and Analytics of Complex Management Systems (CACMS), Tianjin University, Tianjin 30072, China; (e-mail: hongke@tju.edu.cn) 
\vspace{-0.2em}
\IEEEcompsocthanksitem X. Zhou is with the Department of Computer Science and Engineering, The Hong Kong University of Science and Technology, Hong Kong, SAR, China;  (e-mail: zxf@ust.hk). 
}}

\markboth{XXXXXX}%
{Shell \MakeLowercase{\textit{et al.}}: Bare Advanced Demo of IEEEtran.cls for IEEE Computer Society Journals}

\maketitle

\begin{abstract}
 Accurate healthcare prediction is critical for improving patient outcomes and reducing operational costs. 
 Bolstered by growing reasoning capabilities, large language models (LLMs) offer a promising path to enhance healthcare predictions by drawing on their rich parametric knowledge.
 However, LLMs are prone to factual inaccuracies due to limitations in the reliability and coverage of their embedded knowledge. While retrieval-augmented generation (RAG) frameworks, such as GraphRAG and its variants, have been proposed to mitigate these issues by incorporating external knowledge, they face two key challenges {in the healthcare scenario}: (1) identifying the {clinical necessity} to activate the retrieval mechanism, and (2) achieving synergy between the retriever and the generator to craft contextually appropriate retrievals. To address these challenges, we propose GHAR, a \underline{g}enerative \underline{h}ierarchical \underline{a}gentic \underline{R}AG framework that simultaneously resolves when to retrieve and how to optimize the collaboration between submodules in healthcare. Specifically, for the first challenge, we design a dual-agent architecture comprising Agent-Top and Agent-Low. {Agent-Top acts as the primary physician, iteratively deciding whether to rely on parametric knowledge or to initiate retrieval, while Agent-Low acts as the consulting service, summarising all task-relevant knowledge once retrieval was triggered.} To tackle the second challenge, we innovatively unify the optimization of both agents within a formal Markov Decision Process, designing diverse rewards to align their shared goal of accurate prediction while preserving their distinct roles. Extensive experiments on three benchmark datasets across three popular tasks demonstrate our superiority over state-of-the-art baselines, highlighting the potential of hierarchical agentic RAG in advancing healthcare systems.
\end{abstract}





\section{Introduction}\label{sec:intro}
\begin{figure}[!t] 
\centering
\includegraphics[width=\linewidth]{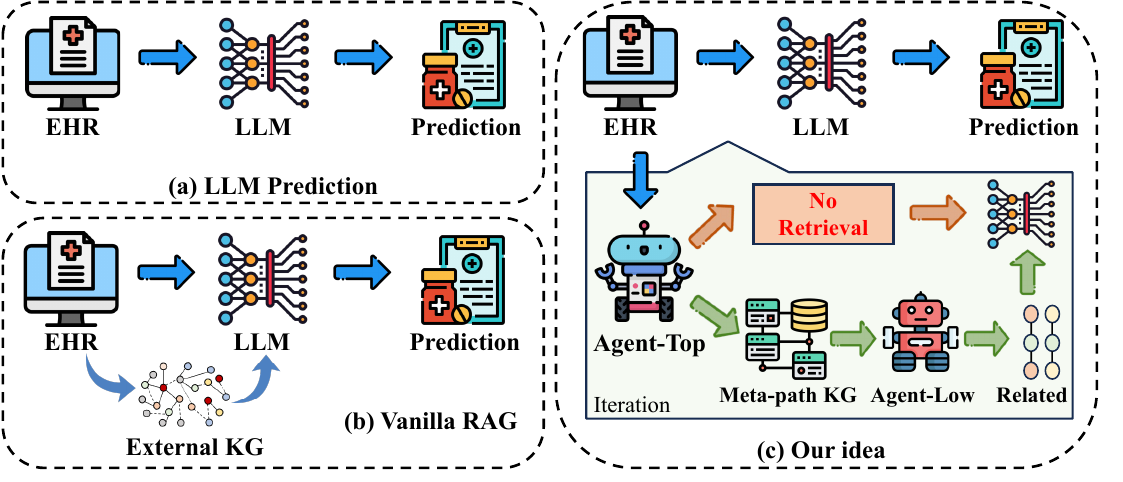}
\setlength{\abovecaptionskip}{-0.05cm}   
\setlength{\belowcaptionskip}{-0.1cm}   
\caption{Motivation Difference. (a) Forecasting using only LLM parameterized knowledge. (b) Single-round retrieve augmented generation with an external knowledge graph (KG). (c) Our idea utilizes hierarchical agents for iterative generation.}\label{fig:motiv}
\vspace{-0.3cm}
\end{figure} 

\IEEEPARstart{H}{ealthcare} is a pivotal domain for societal well-being, fundamentally influencing patient outcomes and the operational effectiveness of medical systems~\cite{zheng2023interaction,zheng2021drug,liu2025generalist}.
Prevailing healthcare prediction approaches can be broadly classified into discriminative and generative paradigms~\cite{chenomi,zang2024colacare}. 
Discriminative approaches, which have predominated in prior research, excel at capturing temporal patterns~\cite{gao2020stagenet} and modeling higher-order co-occurrences among Electronic Health Record (EHR)~\cite{grasp}. 
However, these EHR ID-based methods inherently neglect entity semantics, thereby failing to uncover the underlying mechanisms beyond statistical associations and ultimately limiting model generalizability and interpretability.
Recently, the brilliance of large language models (LLMs) in information retrieval has inspired researchers to explore their applications in healthcare~\cite{jiangkare,zhao2025beyond,improve24}.
Typical approaches involve utilizing LLMs to construct external knowledge bases~\cite{jianggraphcare,zhu2024emerge,xu2025dearllm} or employing them as foundation models for instruction tuning~\cite{bosselut2024meditron,nazar2025design}, aiming to leverage their extensive parametric knowledge to enhance prediction accuracy, as illustrated in Fig.~\ref{fig:motiv}(a).
Despite their potential, LLMs are prone to generating plausible yet factually inaccurate content, a phenomenon known as ``hallucination"~\cite{huang2025survey,mrag24}.  For instance, within the context of medical dialogue, an LLM may erroneously validate a patient's self-diagnosis of ``adrenal fatigue"—a condition not recognized by mainstream endocrinology—by fabricating non-existent diagnostic criteria or citing spurious sources to substantiate its claim.




{
As illustrated in Fig.~\ref{fig:motiv}(b), recent generative initiatives like GraphRAG aim to minimize hallucinations through knowledge-intensive retrieval-augmented generation (RAG), typically employing a single-round retrieval to gather relevant evidence from external sources \cite{graphragsli,guo2024lightrag}. While effective for straightforward tasks, this one-shot approach falls short in healthcare, where composite clinical queries demand intricate reasoning.
Building on this, several RAG-based frameworks have been tailored for medical applications. However, they still encounter two primary limitations.
First, systems like MedRAG~\cite{zhaomedrag} and KARE~\cite{jiangkare} either overlook or utilize static rules to determine the necessity of retrieval, failing to dynamically assess the clinical necessity for additional information. For example, a query related to a common chronic condition such as diabetes may not routinely require additional retrieval, whereas diagnosing a rare genetic disorder could benefit significantly from consulting specialized databases. Over-reliance on unnecessary retrieval may introduce extraneous noise, compromise output quality, and increase inference latency, as demonstrated in Fig.~\ref{fig:case:ret}.
Second, the retriever and the generator—the two core components of the RAG pipeline—often fail to synergize effectively.}
Conventional paradigms treat the training of retrievers and LLM generators as isolated processes~\cite{searchr1,jiangkare}, where the former are typically optimized for document ranking metrics, such as Normalized Discounted Cumulative Gain (NDCG), while the latter are trained for conditional text generation~\cite{lee2025routerretriever}. {This semantic incongruity adversely hampers their collaboration in healthcare. The generator, unaware of the retriever’s selection rationale, tends to process retrieved evidence passively, leading to a ``loss-in-the-middle” phenomenon where crucial clinical cues are obscured by non-actionable information, compromising accuracy.
Consequently, as illustrated in Fig.~\ref{fig:case:case:kg}, while KARE's extracted information is semantically related, it often fails to provide the precise evidence required for reliable clinical decision-making, leading to erroneous predictions. 
}

To address these challenges, we propose GHAR, a novel \underline{g}enerative framework designed to enhance RAG performance in healthcare through a \underline{h}ierarchical \underline{a}gentic \underline{r}etrieval architecture. 
First, to tackle the ``when to retrieve" dilemma, we introduce a dual-agent architecture comprising Agent-Top and Agent-Low, as depicted in Fig.~\ref{fig:motiv}(c). {Agent-Top serves as the primary physician, assessing the need for a deeper workup and initiating additional consultations, while Agent-Low functions as the consulting service that synthesizes the retrieved information into a cohesive summary of relevant clinical knowledge.}
Meanwhile, to achieve a more fine-grained narrowing of the retrieval scope, we introduce the critical concept of meta-path~\cite{zhang2014meta,wang2022survey} within heterogeneous graphs for partition awareness. {Each meta-path (e.g., \textit{disease$\rightarrow$treated\_by$\rightarrow$drug}) represents a distinct type of clinical relationship.} Once retrieval is deemed necessary, Agent-Top selects the appropriate meta-paths for coarse-grained identification of knowledge gaps, {much like a physician identifies primary protocols before conducting a detailed consultation}. {This targeted approach extracts a focused, clinically pertinent data subset, sharpening retrieval precision for medical decision-making, as evidenced in Table~\ref{tab:aba} and Fig.~\ref{fig:time:dis}.}
Second, to unify retriever and LLM generator under a collaborative optimization framework, we formalize the {healthcare decision-making} as a Markov Decision Process (MDP)~\cite{zhaoranm,ching2006markov}, treating each component as a learning agent. We optimize these agents jointly through multi-agent reinforcement learning (RL)~\cite{hu2024review,jiang2024multi}, facilitating coordinated decision-making that enhances overall system performance.
This approach aligns the objectives of LLM reasoning and retrieval, {ensuring they function cohesively to produce accurate and well-supported clinical outputs.} Furthermore, we design a diverse set of rewards tailored to the unique characteristics and commonalities of each agent. Our reward structure emphasizes lower overall costs, higher task accuracy, and the rationality of the reasoning path. Through continuous exploration in optimization, the model progressively identifies optimal strategies, allowing each agent to fulfill its designated role while promoting strong synergy among them.

To summarize, the contributions of this work are threefold:
\vspace{-1em}
\begin{itemize}[leftmargin=12pt]
    \item We are the first to devise a hierarchical agentic RAG tailored for healthcare prediction. GHAR not only resolves the critical issue of ``when to retrieve" but also enables ``collaborative optimization of the generator and retriever", establishing an effective and efficient RAG pipeline.
    \item We innovatively formulate the RAG optimization as a MDP process, employing multi-agent RL to synchronize the reasoning process and partition-aware retrieval within a unified framework. This enables nuanced decision-making, with varied rewards ensuring semantic relevance and synergy among agents.
    \item Extensive experimental results demonstrate the superiority of GHAR over state-of-the-art baselines. 
\end{itemize}

\section{Related Work}\label{sec:rel}
In this section, we review the closely related work, highlighting both connections and distinctions.

\subsection{Healthcare Prediction}\label{sec:rel:health}
Healthcare prediction is a critical component in advancing medical practice, significantly influencing patient outcomes and operational efficiency of healthcare systems~\cite{liu2024moe,zang2024colacare,liu2025generalist}. 

Data-driven healthcare prediction falls into two categories: discriminative and generative~\cite{clinicbench,jiangkare}. Discriminative methods can be further divided into instance-based, longitudinal-based, and hybrid approaches. Instance-based models~\cite{grasp} often formulate graphs to depict relationships among entities in single-round EHRs; however, they are limited by their static nature. Longitudinal methods~\cite{gao2020stagenet,zhao2024enhancing,liu2023shape}, on the other hand, account for sequential patterns and utilize RNN and Transformer in predicting patient outcomes. Despite these advancements, both paradigms remain predominantly ID-centric co-occurrence, prompting the development of hybrid approaches that enhance the utilization of external knowledge. Common strategies~\cite{zhu2024emerge,zhao2025} involve constructing external knowledge graphs or hypergraphs to facilitate the retrieval of relevant subgraphs for each patient visit. Recently, validations of LLMs in information retrieval have prompted researchers to explore their potential for generating additional corpora and supporting generative tasks~\cite{liu2025knowledge}. Their typical approach involves utilizing LLMs' parametric knowledge to enrich external corpora or employing instruction learning to adapt general LLMs to the healthcare domain~\cite{meditron,biomistral}.
Nonetheless, challenges remain, particularly regarding LLM hallucinations, which have not been effectively addressed~\cite{graphragsli}.
Although frameworks like KARE~\cite{jiangkare} and MedRAG~\cite{zhaomedrag} attempt to integrate RAG pipelines for factual verification, their approach suffers from two key limitations. Not only does their indiscriminate reliance on single-round retrieval incur significant costs, but their isolated optimization process also risks creating an alignment gap between the LLM generation and the retrieval process.

\textit{Our approach is grounded in the generative genre. Unlike hybrid discriminative methods, we directly use LLMs as the backbone, allowing us to capture semantic nuances beyond small databases and achieve enhanced decision-making capabilities.
In contrast to the existing generative approaches, we are the first to apply agentic RAG and iterative deep thinking specifically to healthcare prediction, distinguishing our approach from existing single-round RAG models like KARE and MedRAG.  Furthermore, our retrieval process is aligned with overarching healthcare objectives and is optimizable—a feature that is currently lacking in existing work.}

\subsection{Retrieval Augment Generation}\label{sec:rel:rag}
RAG systems mitigate LLM limitations, such as factual hallucinations and outdated knowledge, by retrieving relevant information from external corpora~\cite{fan2024survey,deep25}. This approach is particularly crucial in knowledge-intensive fields like healthcare~\cite{amugongo2024retrieval}, where accuracy and factual evidence are vital for effective clinical decision-making.

The earliest implementation, Naive RAG~\cite{huly2024old}, relies on traditional retrieval techniques such as TF-IDF and BM25 to fetch documents from static datasets. While effective for keyword-based queries, Naive RAG struggles to capture semantic nuances, limiting its contextual understanding. In contrast, Advanced RAG~\cite{karpukhin2020dense,yu2025gar} leverages dense vector search models, like dense passage retrieval, and neural ranking algorithms to encode queries and documents into high-dimensional vector spaces. This enables better semantic alignment and retrieval accuracy, which is crucial in healthcare settings where nuanced understanding can significantly influence patient outcomes. Modular RAG~\cite{modulerrag} further enhances flexibility by decomposing the retrieval and generation pipeline into independent, reusable components, allowing for domain-specific customization tailored to varied healthcare applications.
Building on these concepts, Graph RAG~\cite{li2024glbench,guo2024lightrag,wu2024supporting} incorporates graph-structured data to enhance multi-hop reasoning and contextual richness, making it particularly effective in fields like healthcare and law, where complex relationships between entities are vital. The most recent advancement, Agentic RAG~\cite{agenticrag,searchr1}, introduces autonomous agents capable of dynamic decision-making and workflow optimization. This paradigm employs query refinement and adaptive retrieval strategies to address complex, real-time, multi-domain queries. 

\textit{Our work synthesizes elements from both Graph RAG and Agentic RAG into a cohesive framework. Unlike existing RAG systems that perform retrieval for all queries, our approach employs an iterative reasoning process to determine the optimal timing for retrieving external knowledge. We also introduce meta-paths as innovative partitioning strategies for fine-grained and efficient retrieval. Additionally, we formalize the interactions between agents as a Markov Decision Process (MDP), enabling us to unify the optimization of Agent-Top (which determines when to retrieve) and Agent-Low (which handles task-relevant retrieval) under a shared reward mechanism. Table~\ref{tab:key} outlines the key distinctions between our algorithm and existing generative approaches.
}
\begin{table}[!h]
\centering
\setlength{\abovecaptionskip}{-0.05cm}   
\setlength{\belowcaptionskip}{-0.1cm}   
\caption{Key Difference with generative baselines. All refers to searching in the entire knowledge base, and partially refers to searching in the selected meta-path knowledge partition. NA refers to not applicable. }
\label{tab:key}
\resizebox{0.48\textwidth}{!}{
\begin{tabular}{ccccccc} 
\toprule
Medthod     & External Knowledge & Knowledge Used & Training & LLM-Retriever Optimization & Iteration    & Reward   \\ 
\hline
LightRAG~\cite{guo2024lightrag}    & KG                 & All            & SFT      & Separated                  & Single-round & NA       \\
Search-R1~\cite{searchr1}   & Document           & All            & RL       & Separated                  & Multi-round  & Single   \\
MedRAG~\cite{zhaomedrag}      & KG                 & All            & SFT      & Separated                  & Single-round & NA       \\
KARE~\cite{jiangkare}        & KG                 & All            & SFT      & Separated                  & Single-round & NA       \\
Medical-SFT~\cite{qwen2.5} & NA                 & NA             & SFT      & NA                         & Single-round & NA       \\
\hline
Ours        & KG                 & Partially, (Meta-path)      & SFT, RL  & Unified                    & Multi-round  & Diverse  \\
\bottomrule
\end{tabular}}
\vspace{-1em}
\end{table}

\section{Proposed Method}\label{sec:med}
We first introduce the preliminaries, then provide an overview of the proposed GHAR, and detail submodules.

\subsection{Preliminaries}\label{sec:med:pre}
\noindent\textbf{Healthcare Dataset:} Each patient's medical history is represented as a sequence of visits, denoted as \(\mathcal{U}^{(i)} = (\mathrm{u}^{(i)}_{1}, \mathrm{u}^{(i)}_{2}, \dots, \mathrm{u}^{(i)}_{\mathcal{N}_{i}})\), where \(i\) identifies the \(i\)-th patient in the set \(\mathcal{U}\) and \(\mathcal{N}_{i}\) indicates the total number of visits. For clarity, we use a single patient as an example and omit the superscript $i$.
Each visit yields multiple views of data, represented as \(\mathrm{u}_{j} = \{\mathrm{d}_{j}, \mathrm{p}_{j}, \mathrm{m}_{j}\}\), where diagnosis (\(\mathrm{d} \in \mathcal{D}\)), procedure (\(\mathrm{p} \in \mathcal{P}\)), and medication (\(\mathrm{m} \in \mathcal{M}\)), each represented as a set respectively. For instance, in the \(j\)-th visit,  ${\mathrm{d}}_{j}$, may include \{Heart Failure, Diabetes\}, indicating that the patient has two serious diseases.

\noindent\textbf{Biomedical Knowledge Graph:} Our KG corpora is represented as a heterogeneous graph \(\mathcal{G} = (\mathcal{V}, \mathcal{E})\), where \(\mathcal{V}\) denotes the set of nodes (entities) and \(\mathcal{E}\) denotes the set of edges (relationships). We further use $\mathcal{\tilde{V}}$ to denote the node type. The relationships are primarily sourced from PrimeKG~\cite{chandak2023building,zhu2024emerge}, a popular and comprehensive knowledge base for healthcare. Its primary data sources include public knowledge graphs (e.g., DBpedia, Wikidata), academic literature, and medical databases, ensuring a rich and accurate knowledge graph.  Additionally, following~\cite{zhang2014meta,wang2022survey}, we define meta-paths as sequences of node types and edges that capture specific relationships, represented as ${o} = (\tilde{v}_{i}, e_{ij}, \tilde{v}_j)$, where \(\tilde{v} \in \mathcal{\tilde{V}}\) and \(e_i \in \mathcal{E}\). The set {$\mathcal{O}_{\text{meta}}$ refers to the universal meta-path set, i.e., $\mathcal{O}_{\text{meta}}=\{o_{1},\cdots,o_{\mathcal{X}}\}$, where $\mathcal{X}$ denotes the total number of meta-paths within PrimeKG.

\noindent\textbf{Parametric Knowledge \& RAG:} Parametric knowledge can be formalized as $\text{llm}(f_{*}(\mathrm{u}_{i}))$, where $f_{*}(\cdot)$ denotes the prompt template. Conversely, RAG can be formalized as 
$\text{rag}(f_{*}(\mathrm{u}_{i}, \mathrm{r}_{i}))$,
where \(\mathrm{r}_{i}\) denotes the retrieved relevant content. The core difference between the two approaches lies in whether the retrieval process is invoked.

\noindent\textbf{Task Formulation:}
Based on the literature~\cite{jiangkare,zhu2024emerge,jianggraphcare}, we present the formal definitions for three common tasks in healthcare prediction. Formally,
\begin{itemize}[leftmargin=12pt]
 \item \textbf{24h-Decompensation (DEC Pred)} focuses on a binary classification task aimed at predicting the likelihood of a patient's decompensation risks. It analyzes \([\mathrm{u}_{1}, ..., \mathrm{u}_{j+1}]\) to determine the decompensation risk within 24 hours, where the label \(\mathbf{y}[\mathrm{u}_{j+1}] \in \mathbb{R}^{1\times 2}\), indicating either yes or no.

 \item \textbf{Readmission Prediction (READ Pred)} involves a binary classification task aimed at forecasting the probability of a patient being readmitted within a certain period after discharge. It strives to predict $\mathrm{y}[\phi(\mathrm{u}_{j+1}) - \phi(\mathrm{u}_{j})] $ using $\{\mathrm{u}_1, \mathrm{u}_2, \ldots, \mathrm{u}_{j}\}$. Here, $\phi(\mathrm{\mathrm{u}}_j)$ signifies the encounter time for visit $\mathrm{u}_j$. The outcome $\mathrm{y}[\phi(\mathrm{u}_{j+1}) - \phi(\mathrm{u}_{j})]$ is assigned 1 if $ \phi(\mathrm{u}_{j+1}) - \phi(\mathrm{u}_{j}) \leq \kappa$, otherwise 0. In this study, $\kappa=15$, aligning with~\cite{jianggraphcare,grasp}.

 \item \textbf{Length-of-stay Prediction (LOS Pred)} is formulated as a multi-class classification problem to categorize the predicted length of stay into intervals. The aim is to forecast the length of ICU stay at time $j+1$, based on EHR data $\{\mathrm{u}_{1}, ..., \mathrm{u}_{j+1}\}$. We set intervals $\mathcal{C}$ into 10 splits following~\cite{jianggraphcare,clinicbench}, i.e., $\mathrm{y}[\mathrm{u}_{j+1}] \in \mathbb{R}^{1\times |\mathcal{C}|}$.
\end{itemize}
Key mathematical symbols of this paper are listed in Table~\ref{tab:math}.

\noindent\textbf{Solution Overview}. Our solution aims to identify the optimal moment to activate the retrieval module and collaboratively optimize sub-modules to craft contextually appropriate retrievals. 
We first define an \textbf{Agent-Top}, which is responsible for high-level decisions regarding whether to engage in deeper reasoning and when to trigger the RAG or the LLM. This decision-making process is crucial, as it determines how efficiently the system can respond to dynamic patient needs. 
Upon invoking RAG, Agent-Top dynamically selects relevant meta-paths and transmits them to \textbf{Agent-Low} for focused attention. These meta-paths greatly narrow the knowledge base and offer a coarse-grained preliminary screening of information.
Agent-Low engages directly with the externally extracted knowledge to generate precise and task-relevant responses, serving as contextual input for the historical reasoning paths in subsequent iterations.
The optimization process for the two agents is structured as a unified MDP, supported by diverse rewards to ensure the distinct roles of each agent while aligning with shared task objectives. This strategy collaboratively optimizes the retrieval and generation processes, fostering a bidirectional demand-aware environment. The methodological overview is illustrated in Fig.~\ref{fig:frame}.

\begin{table}\small
\centering
\setlength{\abovecaptionskip}{-0.05cm}   
\setlength{\belowcaptionskip}{-0.1cm}   
\caption{Mathematical Notations.}
\label{tab:math}
\resizebox{0.48\textwidth}{!}{
\begin{tabular}{c|l} 
\toprule
\textbf{Notations}                                                                                                     & \textbf{Descriptions}                                              \\ 
\hline
$\mathcal{U}  $                                                                                            & any ehr dataset                                          \\
$\mathrm{u}$                                                                                            & patient u's ehr data                                        \\
$\mathcal{D},~\mathcal{P},~\mathcal{M}$   & diagnosis, procedure, and medication set                                    \\
$\mathrm{d},~\mathrm{p},~\mathrm{m} $                                  & multi-hot code of diagnosis, procedure, and medication                    \\
$\mathcal{G}$                                                                                            & biomedical KG            \\
$\mathcal{V},\mathcal{E}$                                                                                            & node set, edge set         \\
$\mathcal{\tilde{V}},\mathcal{\tilde{E}}$                                                                                            & node type, edge type         \\
\hline
$\mathcal{O}_{\text{meta}}, \mathcal{\tilde{O}}_{\text{meta}}$                                                                                            & all meta-paths, generated meta-paths    \\
$f_{*}(\cdot)$                                                                                            & prompt template     \\
$\mathcal{H}_{\text{rea}}$                                                                                            & accumulated reason history     \\
\hline
$\mathbf{S},\mathbf{A}, \mathbf{P},\mathbf{R}$                                                                                            & state set, action set, transition probability matrix, rewards \\
                                                                    
$\mathrm{q}, \mathcal{Q}_{\text{sub}}$                                                           & query, query queue  \\ 
$\text{llm}(\cdot), \text{rag}(\cdot)$                                                           & llm generation, rag generation  \\ 
$\Phi(N)$                                                           & Top-N retrieval  \\ 
$\mathcal{I}$ & the number of max iteration \\

\hline
$\mathrm{y}, \mathrm{\hat{y}}$                                                                                      &  ground-truth, prediction                              \\
$\eta$                                                                                  &  trade-off weight                                 \\
$\mathbb{E}$                                                                                      &   mathematics expectations                          \\ 
$\mathbb{I}$                                                                                      &   format examination                          \\ 

$A$                                                                                   &  advantage                        \\ 
$\theta,\psi$                                                                                   &  actor, critic parameters                   \\  
\bottomrule
\end{tabular}}
\vspace{-1em}
\end{table}
\begin{figure*}[!h]
\centering
\includegraphics[width=0.92\linewidth,height=0.32\linewidth]{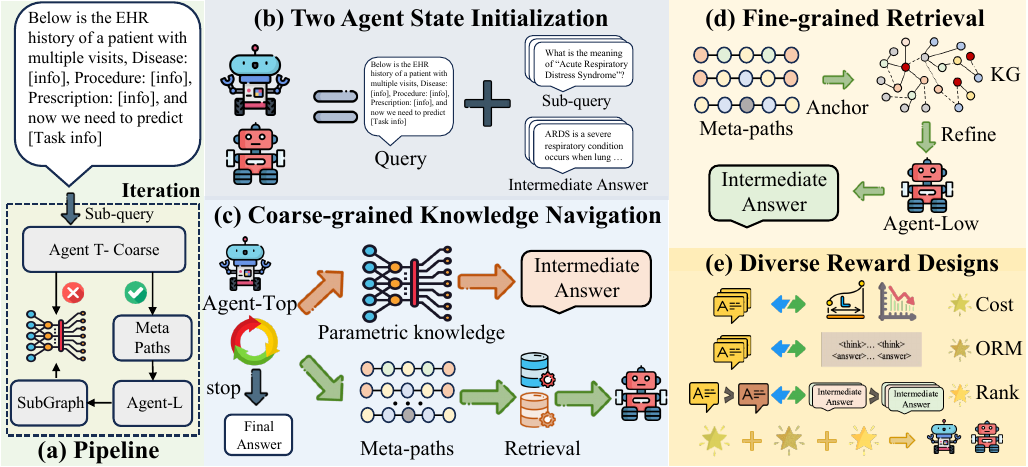}
\centering
\caption{Overview of \textit{GHAR}. (a) Outline of the pipeline for each iteration, potentially involving LLM or LLM+RAG paths. (b) The agent's state is determined by the initial query and all historical reasoning paths. (c) For Agent-Top, it is essential to determine both whether to terminate the process and when to trigger retrieval. (d) For Agent-Low, it summarizes extracted external knowledge to produce a task-relevant response. (e) The diverse rewards design includes cost reduction, format standardization, and accuracy, as well as ranking, to maintain the role distinction and collaborative dynamics between the two agents. ORM denotes the outcome-supervised reward~\cite{deepseek}.}
\label{fig:frame}
\vspace{-0.5cm}
\end{figure*}

\subsection{Overview of the MDP Modeling}\label{sec:med:over}
We first overview the MDP~\cite{zhaoranm,ching2006markov} for the construction of the Agent-Top and Agent-Low. This MDP framework captures the interactions and decision-making processes of both agents. Formally, 
\begin{equation}\label{eq:1}
    \mathbf{MDP} = \left\langle \mathbf{S}^{T} \times \mathbf{S}^{L}, \mathbf{A}^{T} \times \mathbf{A}^{L}, \mathbf{P}, \mathbf{R}^{T} \oplus \mathbf{R}^{L} \right\rangle,
\end{equation}
   where \(\mathbf{S}^{T}\) and \(\mathbf{S}^{L}\) are the sets of states for Agent-Top and Agent-Low, respectively.  \(\mathbf{A}^{T}\) and \(\mathbf{A}^{L}\) denote their action sets, while \(\mathbf{R}^{T}\) and \(\mathbf{R}^{L}\) represent their rewards. $\mathbf{P}$ denotes the transition probability matrix. Then, we present the state transition for the unified MDP model. Formally, for the state of $t+1$-th iteration,
\begin{equation}\label{eq:2}
\begin{aligned}
    s_{t+1}=\left(s_{t+1}^T, s_{t+1}^L\right) &\sim \mathbf{P}\left(\cdot \mid s_t^T, s_t^L, a_t^T, a_t^L\right) \\
    & \sim \mathbf{P}^T(\cdot|s^T_t, a^T_t) \times \mathbf{P}^L(\cdot|s^L_t, a^L_t, \xi(s^T_{t})),
\end{aligned}
\end{equation}
where \(a_t^{T}\) and \(a_t^{L}\) are the actions taken by the Agent-Top and Agent-Low, respectively. $\xi$ denotes hierarchical state transfer into low space. It can be observed that the two agents mutually influence each other, collectively determining the trajectory of the overall state. This interaction provides a rationale for our motivation in pursuing collaborative optimization.
The objective of the MDP is to maximize the expected cumulative reward~\cite{ching2006markov} by selecting an optimal policy \(\pi\), expressed as:
   \begin{equation}\label{eq:3}
   \max_{\pi} \mathbb{E}\left[\sum_{t=0}^{\infty} \gamma^t \left(\mathbf{R}^{T}(s_t, a_t^{T}) + \mathbf{R}^{L}(s_t, a_t^{L})\right)\right],
   \end{equation}
where \(\gamma\) is the discount factor that balances current and future rewards.
The policy \(\pi\) is defined as a mapping from states to actions, determining the best action to take in a given state.

\subsection{Personalized Prompt Collection}\label{sec:med:prompt}
For a healthcare prediction task, we first process the input query based on task requirements and patient information,
\begin{equation}\label{eq:4}
\mathrm{q}^{0}_{\mathrm{u}} = f_{\texttt{query}}(\mathcal{T}, \mathrm{u}),
\end{equation}
where $f_{*}(\cdot)$ denotes the prompt template and $\mathcal{T}$ signifies the task information. Their details can be found in Section~\ref{app:prompt}. Recognizing that a single prompt may introduce bias, we draw on~\cite{lrewrite,lrewrite2} and utilize query rewriting to generate multiple initial queries. Formally,
\begin{equation}\label{eq:5}
\mathrm{Q}_{\text{sub}} = f_{\text{generate}}(\mathrm{q}^0_{\mathrm{u}}, K),
\end{equation}
where $\mathrm{Q}_{\text{sub}}$ denotes the query set and $K$ signifies the number of rewriting. Here we set $K=3$ to maintain prompt diversity. Given that our approach is iterative, we utilize a queue to store $\mathrm{Q}_{\text{sub}}$. 
More precisely, for the $t$-th iteration, we extract the query $\mathrm{q}_{\text{sub}}^{t} \in \mathrm{Q}_{\text{sub}}$ that entered the queue first to proceed with the next agent iteration.

\subsection{Agent-Top: Coarse-grained Knowledge Navigation}\label{sec:med:atop}
Agent-Top acts as a high-level decision-maker, determining whether to engage in deep reflection (i.e., whether to terminate the current iteration) and whether to invoke the partition-aware retrieval process. Its state at iteration $t$ is defined as,
\begin{equation}\label{eq:6}
s_t^T = (\mathrm{q}_{\text{sub}}^{t}, \mathcal{H}_\text{rea}) \in \mathbf{S}^T,
\end{equation}
where $\mathrm{q}_{\text{sub}}^{t}$ is the current sub-query being processed. $\mathcal{H}_\text{rea}$ represents the current accumulated reasoning history. Specifically, for the first subquery, $\mathcal{H}_\text{rea}$ is an empty string `` ", and subsequent entries are updated by appending all previous subqueries and their corresponding intermediate answers, as depicted in Fig.~\ref{fig:frame}(b).
 The action space of Agent-Top includes two-step actions, formally,
\begin{equation}\label{eq:7}
a_t^{T,1} = 
\begin{cases}
\texttt{llm}_{\text{top}}(f_{\text{llm}}(s_{t}^{T})),  \\
\texttt{rag}(f_{\text{meta}}(s_{t}^{T}, \mathcal{O}_{\text{meta}})), 
\end{cases} 
\
a_t^{T,2} = 
\begin{cases}
\text{terminate},  \\
\text{continue},
\end{cases}
\end{equation}
where $a^{T,1} \in \mathbf{A}^{T}$  determines whether to invoke LLM or RAG. If LLM is selected, it directly utilizes its parametric knowledge for response generation; conversely, if RAG is chosen, the process transitions into the RAG framework rag($\cdot$). It is important to note that we do not perform direct retrieval from the entire external KGs. Instead, we focus on generating meta-path IDs $ \mathcal{\tilde{O}}_{\text{meta}}$ and subsequently employ matched meta-paths $\mathcal{\tilde{O}}_{\text{meta}} \cap \mathcal{O}_{\text{meta}}$ for GraphRAG into Agent-Low, facilitating a fine-grained RAG process. This approach mitigates the computational load by constraining semantic neighborhoods within structured relational frameworks, thereby enabling models to focus on semantically relevant subspaces aligned with task-specific meta-paths, as evidenced in Table~\ref{tab:aba} and Fig.~\ref{fig:time:dis}. The parameter \(a^{T,2}  \in \mathbf{A}^{T}\) determines whether to generate subqueries for deeper reasoning. A termination decision indicates that current knowledge suffices for query response. Otherwise, a new subquery will be generated and added to \(\mathrm{Q}_{\text{sub}}\) to bridge the knowledge gap. This simulates the human process of incremental thinking, gradually unraveling to find the final answer to the problem, as detailed in Eq~\ref{eq:13}.
The reward function for Agent-Top is designed to promote reasoning efficiency and path efficacy. Formally,
\begin{equation}\label{eq:8}
\begin{aligned}
\mathbf{R}_{\text{reason}}^{T}&=1-\left|\frac{\operatorname{len}(\mathcal{H}_\text{rea})}{L}-1\right|, \\
\mathbf{R}_{\text{path}}^{T} = |\tilde{\mathcal{O}}_{c}| &- 0.5 \cdot |\tilde{\mathcal{O}}_{e}| - 0.5 \cdot |\tilde{\mathcal{O}}_{r}|,
\end{aligned}
\end{equation}
where $\mathbf{R}_{\text{reason}}^{T}$  evaluates the overall length of the reasoning chain, while $\mathbf{R}_{\text{path}}^{T}$ measures the format of integrated meta-paths. $L$ denotes the expected reason length and is set to 3. $|\tilde{\mathcal{O}}_{c}|, |\tilde{\mathcal{O}}_{e}|, |\tilde{\mathcal{O}}_{r}|$ denote the counts of correct IDs, erroneous IDs, and duplicated IDs within $\mathcal{\tilde{O}}_{\text{meta}}$, respectively. Please note that to ensure the rewards are on the same scale, we follow~\cite{yu2022surprising} for normalization. We encourage shorter chains and effective meta-paths generation, while ensuring performance, as this can reduce the computational burden during the RAG process.

\subsection{Agent-Low: Fine-grained Retrieval}\label{sec:med:alow}
Once Agent-Top decides to trigger the RAG process, the selected meta-paths and subqueries will be transferred to the next stage for processing.
The state of Agent-Low is formulated as,
\begin{equation}\label{eq:10}
s_t^L = (\mathrm{q}_{\text{sub}}^{t},  \mathcal{H}_\text{rea}, \mathcal{G}_{\text{sub}}) \in \mathbf{S}^L,
\end{equation}
where $\mathcal{G}_{\text{sub}}$ is retrieved based on the generated meta-paths in Eq.~\ref{eq:7}. Specifically, we first execute the GraphRAG process for each meta-path partition to extract relevant nodes and edges, thereby constructing the necessary retrieved corpora.
 This subgraph will then be collaboratively passed to Agent-Low to build the state. Formally,
\begin{equation}\label{eq:9}
 \mathcal{G}_{\text{sub}} = \bigcup^{i}_{\mathcal{\tilde{O}}_{c}}\{\Phi_{\text{nodes}}(\mathrm{q}_{\text{sub}}^{t}, \mathcal{G}_{i},N) \cup \Phi_{\text{edges}}(\mathrm{q}_{\text{sub}}^{t}, \mathcal{G}_{i},N)\},
\end{equation}
where $\mathcal{G}_{\text{sub}}$ denotes the retrieved corpus, $\mathcal{G}_{i}$ denotes the KG subgraph under meta-path $i$, and $\Phi(\cdot)$ refers to a specific retrieval algorithm. N refers to the Top-N retrieval in GraphRAG. Here, we choose E5~\cite{e5} as the retriever $\Phi(\cdot)$, and we also present modules for other retrieval algorithms in section~\ref{sec:back}, which are flexible.
Then, the action of Agent-Low is to further summarize these KG subgraphs, generating the task-relevant response as the intermediate answer for the RAG phase. Formally,
\begin{equation}\label{eq:11}
a_t^L = \text{llm}_{\text{low}}(f_{\text{rag}}(s_{t}^{L})) \in \mathbf{A}^{L}, 
\end{equation}
 where $\text{llm}_{\text{low}}$ shares the same LLM with $\text{llm}_{\text{top}}$ for convenience and to ensure semantic consistency.
The reward function for Agent-Low incentivizes concise and relevant retrievals,
\begin{equation}\label{eq:12}
    \mathbf{R}_{\text{rel}}^{L}=\text{Sim}(a_t^L, \mathrm{q}_{\text{sub}}^{t}) + \text{Sim}(a_t^L, \mathcal{G}_{\text{sub}}),
\end{equation}
where $\mathbf{R}_{\text{rel}}$ evaluates how well the retrieved information addresses the sub-query, while also minimizing deviation from the original evidence. Sim($\cdot$) denotes the normalized token overlap count, a widely adopted metric in existing LLM research~\cite{jiang2024multi,hu2024review}.

\subsection{Prediction \& MDP Optimization}\label{sec:med:opt}
\noindent\textbf{Iterative Reasoning.} We adopt an iterative approach to deepen the reasoning process. Specifically, after each subquery receives an intermediate answer, Agent-Top will decide whether to directly use the current information to generate the ultimate answer $\mathrm{\hat{y}}$, as determined by \( a^{T,2} \) in Eq. 2, or to formulate a deeper subquery to bridge the knowledge gap.  Formally,
\begin{equation}\label{eq:13}
\mathrm{\hat{y}} = \text{llm}_{\text{top}}(\mathrm{q}_{\mathrm{u}}^{0}, \mathcal{H}_\text{rea}), \ \ \text{if} \ a^{T,2}=\text{terminate}.
\end{equation}
If $a^{T,2}$ is continue, $\mathrm{Q}_{\text{sub}} \leftarrow \mathrm{Q}_{\text{sub}} \cup \text{llm}_{\text{top}}(f_{\text{sub}}(\mathrm{q}_{\text{sub}}^{t}, \mathcal{H}_\text{rea}))$, referring to add a new subquery to the queue. If the maximum number of iterations $\mathcal{I}$ is reached, it will automatically return ``terminate".
According to this paradigm, the Agent-Top can adaptively determine whether deep thinking is necessary and decide when to terminate. Additionally, the queue data structure facilitates the model's comprehensive understanding of the essential knowledge required to address the problem, dynamically expanding the boundaries of its knowledge, as evidenced in Fig.~\ref{fig:case:case:dec}.

\noindent\textbf{Rewards Design \& Optimization.}
In light of the limitations posed by the independence of previous optimizations for RAG and LLM, we unify the two agents into a single MDP process and optimize them using multi-agent PPO~\cite{yu2022surprising}. To ensure that both agents work collaboratively, we design a shared reward mechanism. Beyond the agent-specific rewards in Eq.~\ref{eq:8} and~\ref{eq:12}, we introduce two additional rewards $\mathbf{R}_{\text{orm}}$ and $\mathbf{R}_{\text{rank}}$ that require collaboration between the agents. Formally, 
\begin{equation}\label{eq:14}
\mathbf{R}_{\text{all}}(s_t, a_t) = \mathbf{R}_{\text{cost}} + \eta\mathbf{R}_{\text{orm}} + \mathbf{R}_{\text{rank}},
\end{equation}
\begin{equation}\label{eq:15}
    \begin{cases}
\mathbf{R}_{\text{cost}}=\mathbf{R}_{\text{reason}}^{T} + \mathbf{R}_{\text{path}}^{T} + \mathbf{R}_{\text{rel}}^{L},\\
        \mathbf{R}_{\text{orm}}= \mathbb{I}(\mathrm{y},\mathrm{\hat{y}}) + \mathbb{I}(\text{format}, \mathrm{\hat{y}}) + \mathbb{I}(\text{format}, {{a}_{t}}),\\
        \mathbf{R}_{\text{rank}}=\max(\alpha, \text{Sim}(\mathcal{H}_\text{rea}, \mathcal{H}_\text{rea}^{\text{pos}}) - \text{Sim}(\mathcal{H}_\text{rea}, \mathcal{H}_\text{rea}^{\text{neg}}))
,\\
    \end{cases}
\end{equation}
where $\mathbf{R}_{\text{orm}}$ denotes the degree of accuracy and format matching with the final answer, i.e., outcome-supervised reward~\cite{deepseek}. We provide additional weight control $\eta$ to balance the trade-off between $\mathbf{R}_{\text{orm}}$ and other auxiliary rewards. $\mathbf{R}_{\text{rank}}$ is used to maintain the distance $\alpha$ between the correct reasoning paths and the negative reasoning paths. 
$\mathcal{H}_\text{rea}^{\text{*}}$ refers to the exploration reasoning path used as a reference.
More specifically, prior to training, we follow prior work~\cite{deepseek,hu2024review,deep25} in using a stronger LLM (Qwen2.5-7B)~\cite{qwen2.5} to generate reasoning rollouts. The common practice is to select correctly answered queries and use their reasoning paths as positive examples ($\mathcal{H}_{\text{rea}}^{\text{pos}}$) for alignment. Unlike them, our approach diverges by additionally recording the model's opposite decision points (e.g., the choice of whether to perform retrieval) and incorporating them into a negative set ($\mathcal{H}_{\text{rea}}^{\text{neg}}$), thereby constructing a pairwise loss. We aim to simultaneously imitate the reasoning process of the correct paths and steer away from the incorrect reasoning paths.

In common multi-agent optimization scenarios, the actor-critic architecture~\cite{zhao2018deep,yu2022surprising} is widely employed, taking into account the reference model($\theta_{\text{ref}}$), actor model($\theta$), and critic model($\psi$). Both agents share parameters using Qwen2.5-3B as the backbone for efficiency, with further possibilities discussed in section~\ref{sec:back}. The designed multi-agent PPO loss is as follows,
\begin{equation}\label{eq:16}
\mathcal{L}_{\text{actor}}(\theta) = \mathbb{E}_{(s_t, a_t) \sim \mathcal{D}} \left[ \min\left(r_{\theta} A_t, \text{clip}\left(r_{\theta}, 1-\epsilon, 1+\epsilon\right) A_t \right) \right],
\end{equation}
where $r_{\theta}=\frac{\pi_\theta(a_t^T, a_t^L \mid s_t)}{\pi_{\theta_{\text{ref}}}(a_t^T, a_t^L \mid s_t)}$ and $\pi_{\theta_{\text{ref}}}$ is a reference policy. $A_t=\sum_{t=0}^{\infty}(\gamma \lambda)^l \delta_{t+l}^V$ denotes the advantage function, where $\delta_{t+l}$ refer to the TD-error~\cite{yu2022surprising}. Correspondingly, we present the optimization for the critic network. Formally,
\begin{equation}\label{eq:17}
 \mathcal{L}_{\text{critic}}(\psi) = \sum_t \left[ (V_\psi(s_t) - \mathbf{R}_{\text{shared}}(s_t, a_t))^2 \right],
\end{equation}
where $V$ is the value estimation and $\mathcal{L}_{\text{critic}}(\psi)$ is used to ensure the stability of the critic network within the actor-critic structure.
Finally, our complete multi-agent optimization loss combines actor loss and critic loss. Formally,
\begin{equation}\label{eq:18}
    \mathcal{L}_{\text{multi-agent}}(\theta, \psi) = \mathcal{L}_{\text{actor}}(\theta) + \mathcal{L}_{\text{critic}}(\psi).
\end{equation}
Our approach significantly differs from previous work that optimizes the LLM generation and retrieval process in isolation.
This unified optimization approach ensures that both agents learn to collaborate effectively while maintaining their specialized roles in the healthcare prediction framework. The overall pipeline of our proposed GHAR can be seen in Algorithm~\ref{alg1}.
\begin{algorithm}\small 
\caption{The Algorithm of \textit{GHAR}.} 
\label{alg1} 
\begin{algorithmic}[1] 
\REQUIRE EHRs $\mathcal{U}$, KG $\mathcal{G}$, Queue $\mathrm{Q}_{\text{sub}}$, Max iteration $\mathcal{I}$;
\ENSURE Policy parameters $\theta$;
\STATE Personalized prompt generation in Eq.~\ref{eq:4};
\STATE Prompt rewriting in Eq.~\ref{eq:5};
\WHILE {$\mathrm{Q}_{\text{sub}}\neq \emptyset \ \& \ t\leq \mathcal{I}$}
    \STATE Extract a query $\mathrm{q}_{\text{sub}}$ from $\mathrm{Q}_{\text{sub}}$;
    \STATE Agent-Top initialization in Eq.~\ref{eq:6};
    \IF {$a_{t}^{T,1}=\text{llm}$} 
        \STATE $a_{t}^{T,1}=\texttt{llm}_{\text{top}}(f_{\text{llm}}(s_{t}^{T}))$;
    \ELSIF {$\ a_{t}^{T,1}=\text{rag}$}
        \STATE \# Agent-Top Meta-path Navigation
        \STATE $a_{t}^{T,1}=\texttt{rag}(f_{\text{meta}}(s_{t}^{T}, \mathcal{O}_{\text{meta}}))$; 
    \STATE \# Agent-Low RAG Process
    \STATE Agent-Low initialization in Eq.~\ref{eq:9};
    \STATE Agent-Low action in Eq.~\ref{eq:11};
    \ENDIF
    \STATE Termination examination in Eq.~\ref{eq:7} \& Eq.~\ref{eq:13};
    \STATE Reward calculation in Eq.~\ref{eq:14}-\ref{eq:15};
    \STATE Actor-critic loss in Eq.~\ref{eq:18};
    \STATE Update the parameters $\theta$, $\psi$;
    \STATE $t=t+1$;
\ENDWHILE \\
\RETURN{} Policy parameters $\theta$
\end{algorithmic}
\end{algorithm}
\vspace{-1em}
\section{Experiments}\label{sec:exp}
\begin{table}\small
\centering
\setlength{\abovecaptionskip}{-0.05cm}   
\caption{Data Statistics. Due to the diverse time requirements of various tasks, we use the LOS Pred task as a representative example, given its lenient data restrictions. \# denotes the number of items, and avg represents the average value. Following the popular data partitioning protocols~\cite{zhao2025,pyhealth}, we divide the dataset into training, validation, and test sets using a 6:2:2 ratio.}
\label{tab:sta}
\resizebox{0.5\textwidth}{!}{
\begin{tabular}{l|ccc} 
\toprule
\textbf{Items}                  & \textbf{MIMIC-III} & \textbf{MIMIC-IV} & \textbf{eICU}  \\ 
\hline
\# of patients / \# of visits   &       35,707 / 44,399            &   46,178 / 154,962   &   8,600 / 18,691              \\
{diagnose. / procedure. / medication.} set size &   6,662 / 1,978 / 197  &      19,438 / 10,790  / 200               & 1,358 / 437  / 1,411  \\
avg. \# of visits         &             1.2434    &  3.3551                &   2.1734 \\
avg. \# of {diagnose} history per visit &          17.7373    &  48.9516               &   14.1709 \\
avg.  \# of {procedure} history per visit &            6.1718    &   8.7626               &   45.9671 \\
avg. \# of {medication} history per visit &              38.2772     &  82.7437                &   26.0229 \\
\bottomrule
\end{tabular}
}
\vspace{-1em}
\end{table}

\noindent\textbf{Datasets \& Baselines.} 
We utilize three popular datasets for healthcare prediction: MIMIC-III~\cite{johnson2019billion}, MIMIC-IV~\cite{johnson2020mimic}, and eICU~\cite{pollard2018eicu}. 
The MIMIC-III database is a large, publicly available dataset of de-identified health data from over 40,000 critical care admissions for research in medical informatics. MIMIC-IV is a more complicated version from the U.S., encompassing over 70,000 admissions and providing more extensive longitudinal EHR histories.
In contrast, the eICU dataset focuses on a diverse cohort of ICU patients across multiple hospitals, encompassing around 200,000 admissions and including detailed clinical data that supports various predictive modeling tasks. We follow~\cite{jiangkare,pyhealth} for data processing, with statistics presented in Table~\ref{tab:sta}.

As detailed in section~\ref{sec:rel:health}, we select fifteen competitive baselines. For discriminative models, we choose approaches such as GRASP~\cite{grasp}, StageNet~\cite{gao2020stagenet}, and SHAPE~\cite{liu2023shape}, all of which are ID-based methods designed to capture the collaborative signals among entities. We also introduce hybrid methods like GraphCare~\cite{jianggraphcare}, EMERGE~\cite{zhu2024emerge}, FlexCare~\cite{xu2024flexcare}, and UDC~\cite{zhao2025}, which enhance traditional discriminative methods using language models (LM-based).
For generative models, we include several tuning-free LLMs, such as DeepSeekR1-7B~\cite{deepseek}, alongside medical LLMs like Meditron-7B~\cite{meditron} and BioMistral-7B~\cite{biomistral}, which are fine-tuned with medical corpora. Several tuning-required baselines from our scenario are also incorporated for comparison.
Medical-SFT (Supervised Fine-Tuning) serves as an effective reasoning method~\cite{qwen2.5} and is employed on the Qwen2.5-7B for a fair comparison.
Additionally, we incorporate RAG methods such as Search-R1\cite{searchr1}, LightRAG~\cite{guo2024lightrag}, KARE~\cite{jiangkare}, and MedRAG~\cite{zhaomedrag}, with the latter two specifically designed for medical contexts. Given that our algorithm utilizes a 7B model during the warm-up phase, we ensure fairness in parametric knowledge by using a consistent LLM backbone for LoRA tuning (rank=8) across all generative models.  For Search-R1 and GHAR, we employ the PPO variant for exploration.
\begin{table*}[!h]\small
\centering
\setlength{\abovecaptionskip}{-0.05cm}   
\setlength{\belowcaptionskip}{-0.1cm}   
\caption{Performance comparison: MIMIC-III dataset. Please note that bold text represents optimal performance. B-Accuracy refers to balanced accuracy~\cite{pyhealth}. The symbol $\uparrow$ indicates that a higher value is better.}
\label{tab:pic}
\resizebox{0.75\textwidth}{!}{
\begin{tabular}{l|c|c|c||c|c||c|c} 
\toprule
\multirow{2}{*}{Genre}            & Task                                        & \multicolumn{2}{c||}{24h-Decompensation}                                                                                         & \multicolumn{2}{c||}{Readmission Prediction}                                                                                      & \multicolumn{2}{c}{Length-of-Stay Prediction}                                                                                                                                  \\ 
\cline{2-8}
                                  & Method                                                   & Accuracy$\uparrow$                                                                       & B-Accuracy$\uparrow$                                           & Accuracy$\uparrow$                                                                       & B-Accuracy$\uparrow$                                     & Accuracy$\uparrow$                                  & F1-score$\uparrow$                                                                                                         \\ 
\hline
\multirow{4}{*}{ID-based}         & GRASP~\cite{grasp}     &   0.9803           & 0.6858                                                                                                                           & 0.5727     & 0.5710                                                                                              &    0.4046                                    & 0.3271                                                                                                                                                                                          \\
                                  & StageNet~\cite{gao2020stagenet}                        &  0.9779      & 0.7109                                                                                                              & 0.5938 & 0.5911                                                                                                                        & 0.4091                                        & 0.3715                                                                                                                     \\
                                  & SHAPE~\cite{liu2023shape}                                    &  0.9769 &    0.8133                                                                                                         &   0.6009    & 0.6074                                                                                                                             & 0.4245                                          & 0.3891                                                                                                               \\ 
\hline
\multirow{4}{*}{LM-based}      & GraphCare~\cite{jianggraphcare}                               & 0.9826   &   0.7675                                                                                                                 & 0.6084    &    0.6069                                                                                                                          &    0.4212                                       &  0.3918 \\
                                  & EMERGE~\cite{zhu2024emerge}                                  &  0.9804   &    0.7770                                                                                                         &   0.6144 &   0.6104                                                                                                                       &  0.4189                                         & 0.3912                                                                                                                   \\
                                  & FlexCare~\cite{xu2024flexcare}                              &0.9816    &   0.8146                                                                                 &  0.6185   &  0.6107                                                                                                                                                               & 0.4320                                          & 0.3937                                                                                                                     \\
                                  & UDC~\cite{zhao2025}                                     & 0.9828  &  0.8166                                                                                                                     & 0.6139    &  0.6188                                                                                                                           & 0.4357                                          &  0.3970                                                                                                                 \\ 
\hline
\multirow{8}{*}{Generative-based} 
                                  & DeepSeekR1-7B~\cite{deepseek}                             &  0.1604   &  0.5379                                                                                                                         & 0.5307  &  0.4906                                                                                                                             &  0.1527                                         &  0.1366                                                                                                                         \\
                                  
                                  & Meditron-7B~\cite{meditron}                                                                        & 0.2581    & 0.4813                                                                                                                &  0.5376      & 0.4932                                                                                      & 0.0322                                          &  0.0141                                                                                                                       \\
                                  &BioMistral-7B~\cite{biomistral}                                                                      & 0.1078    & 0.5280                                                                                & 0.5463    &  0.5043                                                                                                                            & 0.1109                                          & 0.0667                                                                                                                          \\
\cline{2-8}
                                   & LightRAG~\cite{guo2024lightrag}                 & 0.7721   & 0.4836                                                  & 0.5564    & 0.5038                                                                                   &  0.2860                                         & 0.1860                                                                                                                     \\

                                & Search-R1\cite{searchr1}                    & 0.9648  & 0.5561                                     &  0.5314  &   0.5117                                                                                 &   0.3142                                   &   0.2583                                                                                                              \\ 

                        & Medical-SFT~\cite{qwen2.5}                          & 0.9672   & 0.8145                                                                              & 0.5626   &  0.5279                                                                                                                          &  0.4299                                         &   0.3957                                                                               \\
                                
                    & MedRAG~\cite{zhaomedrag}                     & 0.9829  &  0.4984              &  0.5725                                         &  0.5269                                                             &  0.2312                                         & 0.2045                                                                                                            \\ 
                                  & KARE~\cite{jiangkare}                    & 0.9760  & 0.8114              &   0.5769                         & 0.5156                                                                                   &  0.3700                                         & 0.3427                                                                                               \\

\hhline{~---||--||--}
                                  & {\cellcolor[rgb]{0.8,0.8,0.8}}\textit{GHAR} & {\cellcolor[rgb]{0.8,0.8,0.8}}\textbf{0.9842} & {\cellcolor[rgb]{0.8,0.8,0.8}}\textbf{0.8421} & {\cellcolor[rgb]{0.8,0.8,0.8}}\textbf{0.6244} & {\cellcolor[rgb]{0.8,0.8,0.8}}\textbf{0.6272} & {\cellcolor[rgb]{0.8,0.8,0.8}}\textbf{0.4486} & {\cellcolor[rgb]{0.8,0.8,0.8}}\textbf{0.4124} \\
\bottomrule
\end{tabular}}
\end{table*}

\begin{table*}[!h]\small
\centering
\setlength{\abovecaptionskip}{-0.05cm}   
\setlength{\belowcaptionskip}{-0.1cm}   
\caption{Performance comparison: MIMIC-IV dataset.}
\label{tab:miv}
\resizebox{0.75\textwidth}{!}{
\begin{tabular}{l|c|c|c||c|c||c|c} 
\toprule
\multirow{2}{*}{Genre}            & Task                                        & \multicolumn{2}{c||}{24h-Decompensation}                                                                                         & \multicolumn{2}{c||}{Readmission Prediction}                                                                                      & \multicolumn{2}{c}{Length-of-Stay Prediction}                                                                                                                                  \\ 
\cline{2-8}
                                  & Method                                                    & Accuracy$\uparrow$                                                                       & B-Accuracy$\uparrow$                                           & Accuracy$\uparrow$                                                                       & B-Accuracy$\uparrow$                                     & Accuracy$\uparrow$                                  & F1-score$\uparrow$                                                                                                           \\ 
\hline
\multirow{4}{*}{ID-based}         & GRASP~\cite{grasp}                                    &  0.9970     & 0.5995                                                                                                        & 0.6261  &  0.6263                                                                                                                      & 0.3493                                        & 0.3273                                                                                                  \\
                                  & StageNet~\cite{gao2020stagenet}                               & 0.9911    &  0.6532                                                                                                                      &   0.6324          & 0.6333                                                                                                                          &  0.3467                                         &  0.3253                                                                                                         \\
                                  & SHAPE~\cite{liu2023shape}                                 &0.9968       &     0.6798                                                                                                                    & 0.6325   &   0.6371                                                                                                                         & 0.3751                                          & 0.3598                                                                                                   \\ 
\hline
\multirow{4}{*}{LM-based}      & GraphCare~\cite{jianggraphcare}                              & 0.9968    &  0.7030                                                                                                               & 0.6327         & 0.6403                                                                                                                             &   0.3735                                        & 0.3485                                                                                                              \\
                                  & EMERGE~\cite{zhu2024emerge}                                  &  0.9962   &      0.6976                                                                                                                   & 0.6316   &  0.6399                                                                                                                            & 0.3859                                          &  0.3554                                                                                                           \\
                                  & FlexCare~\cite{xu2024flexcare}                               & 0.9960    &   0.7256                                                                                                                     & 0.6282    &   0.6354                                                                                                                         &  0.3658                                         &   0.3348                                                                                                 \\
                                  & UDC~\cite{zhao2025}                                &  \textbf{0.9971}    &    0.7463                                                                                                                   & 0.6388 & 0.6431                                                                                                                            &  0.3916                                         &    0.3544                                                                                              \\ 
\hline
\multirow{8}{*}{Generative-based} 
                                  & DeepSeekR1-7B~\cite{deepseek}            &  0.1421                                                             &   0.5436                                                                                                             & 0.5081         &  0.5026                                                                                  & 0.0788                                          & 0.0510                                                                                             \\
                                  & Meditron-7B~\cite{meditron}                                                                      & 0.1181  &  0.4848                                                                                                                   & 0.5089     &  0.5025                                                                                  & 0.0724                                          & 0.0245                                                                                                              \\
                                  &  BioMistral-7B~\cite{biomistral}                                                                     &  0.1225    &   0.4937                                                                                & 0.5092 &  0.5035                                                                                                                          &  0.0814                                         & 0.0239                                                                                                             \\
                                  \cline{2-8}

                                &LightRAG~\cite{guo2024lightrag}                & 0.2582                                                   & 0.6283                                                                                                                   & 0.5040    & 0.4978                                                                                   & 0.2129                                          &  0.1856                                     \\
                            & Search-R1\cite{searchr1}                                                     &  0.9586           &   0.5295                                                                                                                  & 0.5038  &  0.6020                                                                                  &  0.2706                                         &  0.2819                                                                \\  
                            & Medical-SFT~\cite{qwen2.5}              &   0.9932                                                         & 0.8110                                                                                                                  & 0.6320     & 0.6335                                                                                   &    0.3869                                       & 0.3564                                                 \\

                                  & MedRAG~\cite{zhaomedrag}                                                        & 0.9960                &  0.5705                                                                                                                    &  0.5868                                         &  0.5683                                         &  0.2623                                         &   0.1701                                             \\ 
    
                            & KARE~\cite{jiangkare}                                                 &    0.9851                & 0.7978                                                                                                             & 0.6258     &  0.6446                                                                                 &  0.3274                                         &    0.2865                                                                 \\

                                  
\hhline{~---||--||--}
                                  & {\cellcolor[rgb]{0.8,0.8,0.8}}\textit{GHAR}  & {\cellcolor[rgb]{0.8,0.8,0.8}}{0.9939} & {\cellcolor[rgb]{0.8,0.8,0.8}}\textbf{0.8279} & {\cellcolor[rgb]{0.8,0.8,0.8}}\textbf{0.6473} & {\cellcolor[rgb]{0.8,0.8,0.8}}\textbf{0.6484} & {\cellcolor[rgb]{0.8,0.8,0.8}}\textbf{0.4035} & {\cellcolor[rgb]{0.8,0.8,0.8}}\textbf{0.3678}  \\
\bottomrule
\end{tabular}}
\end{table*}

\begin{table*}[!h]\small
\centering
\setlength{\abovecaptionskip}{-0.05cm}   
\setlength{\belowcaptionskip}{-0.1cm}   
\caption{Performance comparison: eICU dataset.}
\label{tab:eicu}
\resizebox{0.75\textwidth}{!}{
\begin{tabular}{l|c|c|c||c|c||c|c} 
\toprule
\multirow{2}{*}{Genre}            & Task                                        & \multicolumn{2}{c||}{24h-Decompensation}                                                                                           & \multicolumn{2}{c||}{Readmission Prediction}                                                                                      & \multicolumn{2}{c}{Length-of-Stay Prediction}                                                                                                                                  \\ 
\cline{2-8}
                                  & Method                                                  & Accuracy$\uparrow$                                                                       & B-Accuracy$\uparrow$                                           & Accuracy$\uparrow$                                                                       & B-Accuracy$\uparrow$                                     & Accuracy$\uparrow$                                  & F1-score$\uparrow$                                                                                                 \\ 
\hline
\multirow{4}{*}{ID-based}         & GRASP~\cite{grasp}                                  & 0.9238     & 0.5114                                                                           &   0.8944   & 0.5575                                                                                                                                                       &    0.2424                                       & 0.2200                                                                                         \\
                                  & StageNet~\cite{gao2020stagenet}                            & 0.9699        &     0.5294                                                                                                                  &  0.9119 &  0.5319                                                                                            & 0.2646                                        & 0.2145                                                                                                                                                                                    \\
                                  & SHAPE~\cite{liu2023shape}                            &  0.9633         &   0.5744                                                                                                          &  0.9132       &  0.5452                                                                                                                        &  0.3081                                         & 0.2551                                                                                                                        \\ 
\hline
\multirow{4}{*}{LM-based}      & GraphCare~\cite{jianggraphcare}                              & 0.9712    &   0.5200                                                                          &  0.9026                                         &  0.5501                                                                                                                         & 0.2948                                          &    0.2510                                                                                                       \\
                                  & EMERGE~\cite{zhu2024emerge}                               & 0.9703       &  0.5213                                                                                                                            &   0.9167       &    0.5799                                                                                                           &  0.3088                                         & 0.2570                                                                                                        \\
                                  & FlexCare~\cite{xu2024flexcare}                              & 0.9716      & 0.5279                                                                                                                       &  0.9073   &    0.5420                                                                                                                       & 0.2982                                          & 0.2314                                                                                                       \\
                                  & UDC~\cite{zhao2025}                                  & 0.9711   &  0.5476                                                                                                                 &  0.9169       &  0.5517                                                                                                                       &  0.3091                                         &  0.2595                                                                                    \\ 
\hline
\multirow{8}{*}{Generative-based} 
                                  & DeepSeekR1-7B~\cite{deepseek}                                                                     & 0.1676    & 0.5360                                                                          & 0.6320  & 0.4625                                                                                                                            & 0.0718                                          & 0.0456                                                                                               \\
                                  & Meditron-7B~\cite{meditron}                                                                     & 0.0400     & 0.5037       & 0.9168                                                                           & 0.5027                                                                                                                           &  0.2115                                         & 0.0905                                                                                             \\
                                  &  BioMistral-7B~\cite{biomistral}                          & 0.0698                                               & 0.5157          &  0.8990                                                                  & 0.5057                                                                                                                          & 0.0857                                          & 0.0589                                                                                             \\

    \cline{2-8}
                                   & LightRAG~\cite{guo2024lightrag}                          &  0.9468       &  0.4896                                                            &  0.8990                                                   &   0.4847                                                                              & 0.2243                                          &  0.1045                                                                                                                  \\
  
                                    & Search-R1\cite{searchr1}                         &  0.9533        &   0.5006                                                                             & 0.9082            &  0.4999                                                                              & 0.1907                                          &   0.1186                                                                                                               \\  
                                    & Medical-SFT~\cite{qwen2.5}                             & 0.9678     &   0.5183                                                                          &  0.9089                                     & 0.5009                                                                              &  0.2740                                         &  0.1950                                            \\
                                    
                                   & MedRAG~\cite{zhaomedrag}   & 0.9453                                  & 0.4989                                                                                                            & 0.9163   &  0.5014                                                                                 &  0.2179   &  0.1373                                                                                                                     \\ 
                                  & KARE~\cite{jiangkare}                                &   0.9670    &  0.5178                                                                                                          &  0.9100          & 0.5107                                                                               & 0.2779                                          & 0.1773                                  \\  

\hhline{~---|--||--}
                                  & {\cellcolor[rgb]{0.8,0.8,0.8}}\textit{GHAR} & {\cellcolor[rgb]{0.8,0.8,0.8}}\textbf{0.9720} & {\cellcolor[rgb]{0.8,0.8,0.8}}\textbf{0.5990} & {\cellcolor[rgb]{0.8,0.8,0.8}}\textbf{0.9171} & {\cellcolor[rgb]{0.8,0.8,0.8}}\textbf{0.5908} & {\cellcolor[rgb]{0.8,0.8,0.8}}\textbf{0.3200} & {\cellcolor[rgb]{0.8,0.8,0.8}}\textbf{0.2610}   \\
\bottomrule
\end{tabular}}
\vspace{-0.3cm}
\end{table*}

\noindent\textbf{Implementation Details.}
We implement GHAR and all baselines using PyTorch 3.11 and LangChain~\footnote{https://github.com/langchain-ai/langchain}, with a learning rate and training epochs set to 1e-4 and 3, respectively. We conduct our experiments on a hardware configuration featuring a 12-core Intel Xeon CPU and eight NVIDIA A800 GPUs.  In the initial phase, akin to~\cite{deepseek,hu2024review,deep25}, we first employ Qwen2.5-7B~\cite{qwen2.5} for rejection sampling~\cite{deepseek} on the training set to obtain warm-up samples. More precisely, we leverage the reasoning pathways of correctly answered samples from this more advanced LLM as warm-start data, ensuring that the model effectively learns the initial reasoning ability. Next, we conduct two phases of training: SFT ensures the model learns the basic format, while RL allows for further exploration to enhance performance. We set Agent-Top and Agent-Low to share the same LLM parameters at the start of training, specifically Qwen2.5-3B~\cite{qwen2.5}, to reduce the inference load. We select E5~\cite{e5} as the retriever and utilize FAISS~\cite{johnson2019billion} to create the index.  We also explore additional backbones in section~\ref{sec:back} to test their applicability.
 The max number of meta-paths $|\mathcal{\tilde{O}}_{\text{meta}}|$, Top-N recall $N$, and ORM weights $\eta$ are configured to 3, 1, and 5, respectively, based on the hyperparameter search outlined in section~\ref{sec:rob:hyper}. To enhance our training efficiency, we utilize the Ray~\footnote{https://www.ray.io/} framework for distributed training, as shown in section~\ref{sec:rob:com}. The evaluation metrics selected include B-Accuracy, Accuracy, and F1-score, sourced from~\cite{jianggraphcare,zhong2024meddiffusion,jiangkare}, which demonstrate significant clinical relevance. 

\noindent\textbf{Overall Results.}
As demonstrated in Tables~\ref{tab:pic} to \ref{tab:eicu}, our model outperforms all baseline models across three tasks, particularly in performance metrics that consider both positive and negative instances. From an algorithmic perspective, general-purpose LLMs like DeepSeekR1 exhibit limited performance due to gaps in the training corpus, which hinder their effectiveness in clinical scenarios. Effective supervised Medical-SFT can significantly mitigate this issue; however, the reliance on static datasets in SFT restricts generalization capabilities across diverse contexts. In contrast, RAG's ability to integrate external knowledge sources enhances its contextual awareness and adaptability to various queries, as evidenced by KARE's performance on READ tasks.
We further investigate these distinctions in Section~\ref{sec:rob:ood}, particularly regarding out-of-distribution generalization scenarios, where adaptability is crucial. RAG-based baselines, such as KARE and GHAR, outperform most ID-based baselines on the DEC (MIMIC-IV) task. This advantage arises from KARE and GHAR's capacity to leverage a sophisticated understanding of language and external knowledge, enabling them to capture complex semantic relationships and contextual nuances. In contrast, ID-based methods, such as GRASP, primarily focus on co-occurrence relationships, often neglecting the intricacies of language, which limits their performance.
However, this observation is not universally applicable. In LOS Pred tasks, which are characterized by richer interactions, the performance gap narrows and may even reverse. This is likely due to the presence of more collaborative signals, which can lead to more definitive co-occurrence relationships.

In terms of task complexity, LOS predictions, along with the challenges of multi-class tasks and class imbalance, require a deeper understanding of label interactions and patient status comprehension. LM-based hybrid models, such as GraphCare and FlexCare, experience performance degradation in this context due to their reliance on discriminative approaches that depend on co-occurrence patterns. The noise inherent in frozen embedding processes can also introduce inaccuracies that negatively impact performance. Similarly, MedRAG and KARE, which do not optimize the retrieval process or employ selective retrieval, are likely to encounter similar retrieval noise during their generation process, ultimately leading to loss-in-middle. Moreover, we observe that as the training dataset size increases, the performance of ID-based models gradually improves. For instance, on the MIMIC-IV dataset, ID-based models demonstrate enhanced performance due to a greater number of co-occurrence relationships, which aid in understanding patient states. Models like StageNet and SHAPE exhibit stronger performance, even surpassing Search-R1 and KARE on DEC and LOS tasks. In contrast, the impact on LLM-based baselines is minimal; their performance remains stable, as they rely on a comprehensive understanding of contextual relevance that is not solely dependent on co-occurrence signals. This resilience highlights the advantage of LLMs in managing diverse and complex datasets, enabling them to maintain performance even in smaller training scenarios, such as with the MIMIC-III dataset. We further explore out-of-distribution (OOD) reasoning in Section~\ref{sec:rob:ood} to substantiate this point.

In summary, GHAR demonstrates flexibility in utilizing external knowledge, improving performance across diverse tasks and datasets.

\section{Model Analysis and Robust Testings}\label{sec:rob}
We conduct numerous robustness experiments to provide a more in-depth analysis.  Without loss of generality, we use MIMIC-III for examination.
\begin{table}[!ht]
\centering
\setlength{\belowcaptionskip}{-0.1cm}   
\caption{Ablation Study. NI refers to the use of a single-round rather than an iterative RAG. NT indicates the absence of Agent-Top. NL signifies the exclusion of Agent-Low. NM refers to retrieval performed directly over the entire knowledge graph, as opposed to extraction via meta-path partitioning. NS indicates that no shared reward is utilized, resulting in separate optimization for the two agents.
}
\label{tab:aba}
\resizebox{0.48\textwidth}{!}{
\begin{tabular}{c|c|ccccc|c} 
\hline
Algorithms                & Metric   & -NI & -NT & -NL & -NM & -NS & \textbf{GHAR}  \\ 
\hline
\multirow{1}{*}{DEC Pred} & B-Accuracy    & 0.8295  & 0.8072   & 0.8151   & 0.8338  & 0.8099  & \textbf{0.8421}      \\
\hline
\multirow{1}{*}{READ Pred} & B-Accuracy    & 0.6221     &  0.6008    & 0.6169     & 0.6225     & 0.6096    & \textbf{0.6272}        \\
\hline
\multirow{1}{*}{LOS Pred} & F1-score & 0.3928     & 0.3892     & 0.3804     &  0.4038  & 0.3919     & \textbf{0.4124}        \\
\hline
\end{tabular}}
\vspace{-1em}
\end{table}
\subsection{Ablation Studies}\label{sec:rob:aba}
We conduct extensive ablation analyses on the designed submodules while keeping other components consistent to validate the effectiveness of each element. As shown in Table~\ref{tab:aba}, each submodule is indispensable, as evidenced by the performance decline observed with any variant.
GHAR-NI exhibits a performance degradation of over 2\% in both DEC and LOS prediction tasks; the single-round retrieval lacks the depth of consideration necessary for effective outcomes. More precisely, it simply extracts external information based on the original query, losing the ability for a deep understanding of the knowledge gaps in the reasoning process.
GHAR-NT and GHAR-NL, which respectively ablate Agent-Top and Agent-Low, rely solely on a single-agent role to complete tasks. The former approach degenerates into one that unconditionally pulls information from the entire KG for all queries, incurring a heavy computational burden and introducing noise by adding context even when RAG is not required.
The latter crudely employs coarse-grained retrieval documents, potentially preventing the model from refining task-relevant knowledge, leading to a 4\% performance drop (DEC Pred).
GHAR-NM does not implement meta-path partitioning, directly extracting knowledge from medical entities across the entire knowledge graph.  This straightforward Top-N extraction may be limited to coarse-grained knowledge structures, missing opportunities for finer-grained exploration. GHAR-NS eliminates shared rewards, optimizing the two agents separately.  Our findings indicate that this approach results in a significant performance degradation of nearly 3\%, primarily due to the lack of a unified objective among the agents. This misalignment leads to divergent semantic spaces between the LLM and the RAG components, thereby undermining their collaborative effectiveness.
In summary, these experiments demonstrate the effectiveness of our submodules and provide deeper insights.

\subsection{Diverse Retrievers \& LLMs \& Evaluation Settings}\label{sec:back}
We replace various retrievers and LLM generators to assess the pluggability of our algorithm, as both play indispensable roles within our generative framework~\cite{zang2024colacare,guo2024lightrag}.
\begin{figure}[!h] 
\centering
\subfigure[DEC Pred (B-ACC)]{
\begin{minipage}[t]{0.325\linewidth}
\centering
\includegraphics[width=\linewidth,height=0.75\linewidth]{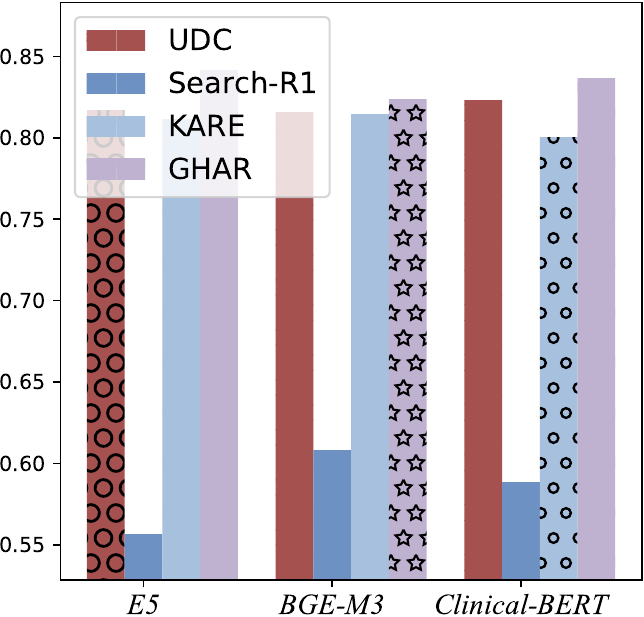}\label{fig:prm:dec:bacc}
\end{minipage}%
}%
\subfigure[READ Pred (B-ACC)]{
\begin{minipage}[t]{0.325\linewidth}
\centering
\includegraphics[width=\linewidth,height=0.75\linewidth]{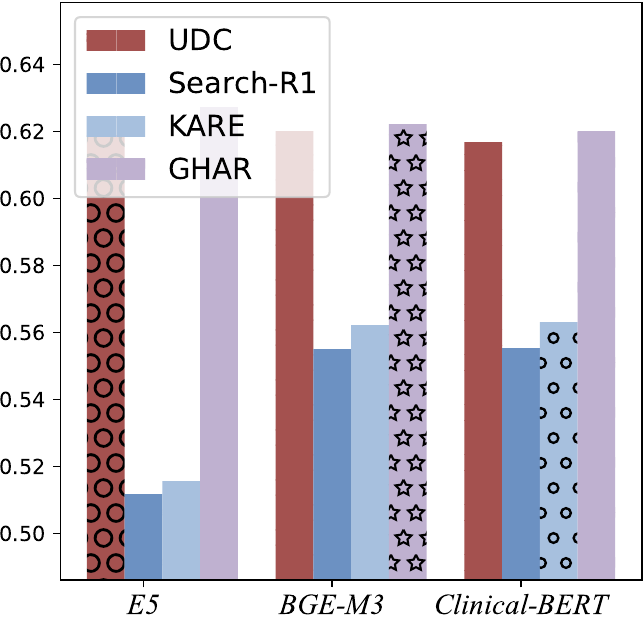}\label{fig:prm:rea:bacc}
\end{minipage}%
}%
\subfigure[LOS Pred (F1-score)]{
\begin{minipage}[t]{0.325\linewidth}
\centering
\includegraphics[width=\linewidth,height=0.75\linewidth]{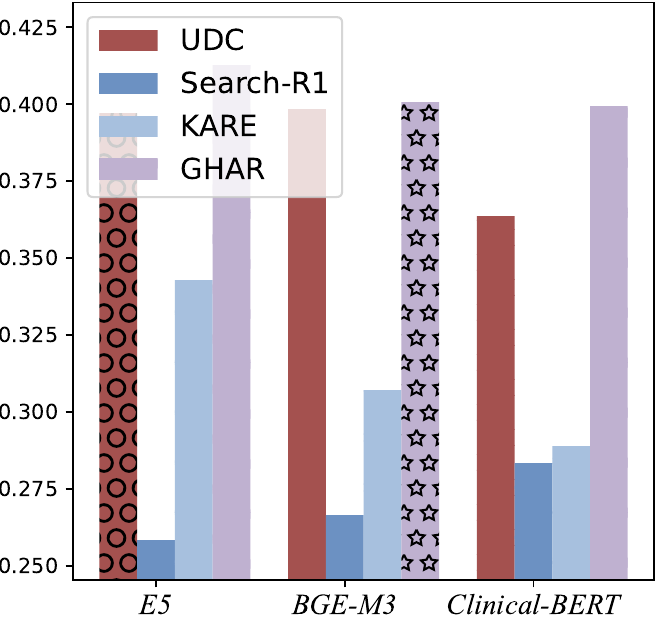}\label{fig:prm:los:f1}
\end{minipage}%
}%
\centering
\setlength{\abovecaptionskip}{-0.15cm}   
\setlength{\belowcaptionskip}{-0.1cm}   
\caption{Comparison under Diverse Retrievers. We employ the popular E5~\cite{e5}, BGE-M3~\cite{bge3}, and Clinical-BERT~\cite{wang2023optimized}.}
\label{fig:plug:prm} 
\vspace{-0.3cm}
\end{figure}
\begin{figure}[!h] 
\centering
\subfigure[DEC Pred (B-ACC)]{
\begin{minipage}[t]{0.325\linewidth}
\centering
\includegraphics[width=\linewidth,height=0.75\linewidth]{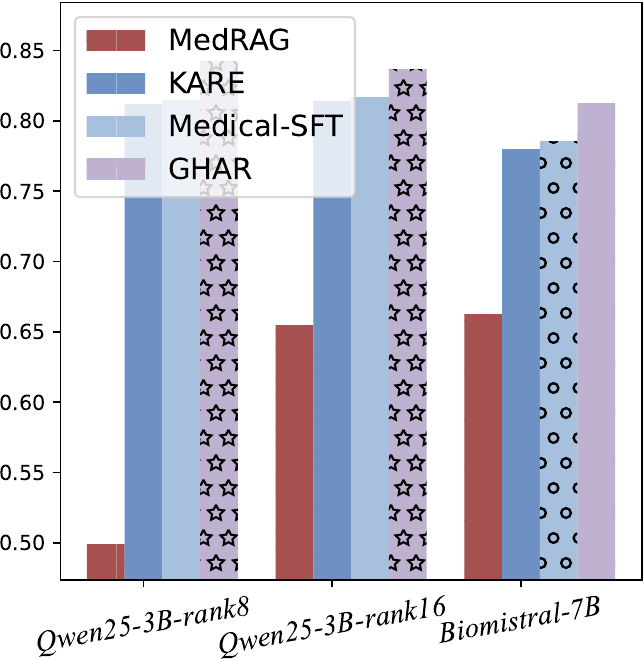}\label{fig:llm:dec:bacc}
\end{minipage}%
}%
\subfigure[READ Pred (B-ACC)]{
\begin{minipage}[t]{0.325\linewidth}
\centering
\includegraphics[width=\linewidth,height=0.75\linewidth]{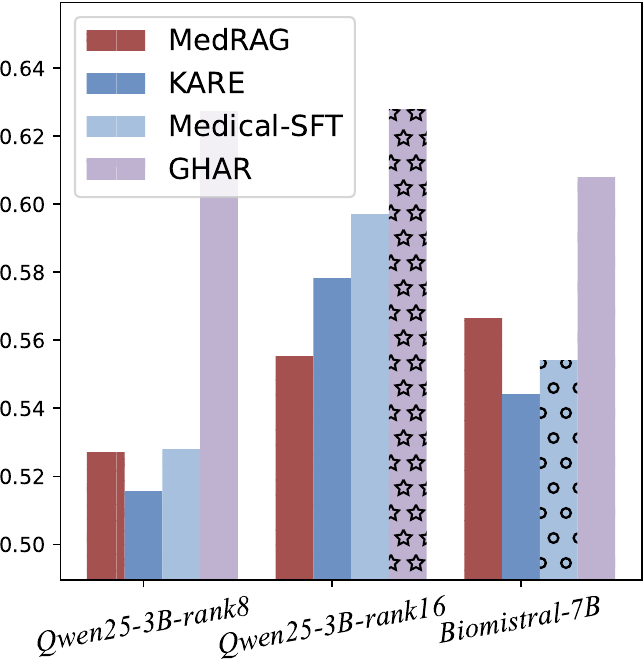}\label{fig:llm:rea:bacc}
\end{minipage}%
}%
\subfigure[LOS Pred (F1-score)]{
\begin{minipage}[t]{0.325\linewidth}
\centering
\includegraphics[width=\linewidth,height=0.75\linewidth]{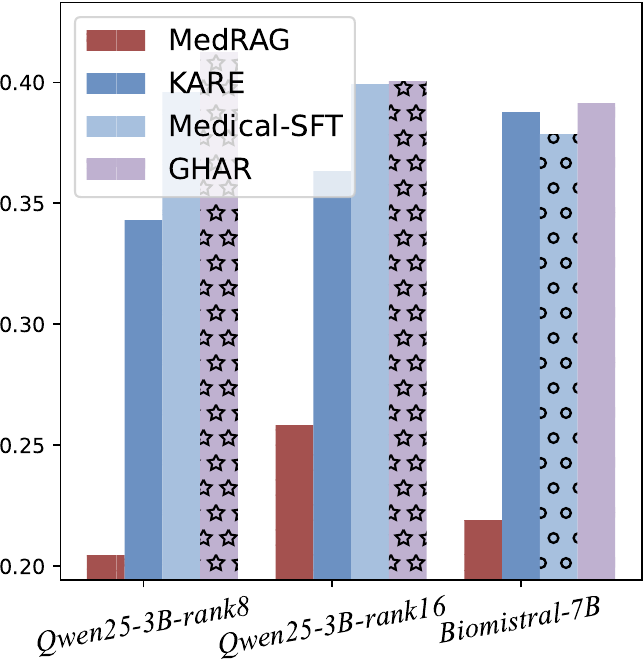}\label{fig:llm:los:f1}
\end{minipage}%
}%
\centering
\setlength{\abovecaptionskip}{-0.15cm}   
\setlength{\belowcaptionskip}{-0.1cm}   
\caption{Comparison under Diverse LLMs. We employ Qwen2.5-3B with rank 8 (our method), Qwen2.5-3B with rank 16~\cite{qwen2.5}, and BioMistral-7B~\cite{biomistral}.}
\label{fig:plug:llm}
\end{figure}

\noindent\textbf{Diverse Retrievers.}
Different retrieval models influence the nodes matched during the extraction of KGs, as variations in embedding spaces arise from distinct pretraining corpora and methodologies~\cite{zhao2025}. As depicted in Fig.~\ref{fig:plug:prm}, BGE-M3~\cite{bge3} and Clinical-BERT~\cite{wang2023optimized} do not demonstrate significant gains. Despite BGE-M3 having more parameters and Clinical-BERT being fine-tuned for medical scenarios, we observe that the content extracted by both models shows only subtle differences. This may be due to significant language discrepancies among nodes during the KG matching process, as well as a gap between their pretraining corpora and healthcare scenarios.
From another perspective, our Agent-Low can refine the retrieval content by focusing on subtle differences, enabling it to obtain information that is closely related to the task. This means it is less influenced by retrieval outputs when the differences between them are minimal.

\noindent\textbf{Diverse LLM Backbones.}
We replace the LLM backbone (Qwen2.5-3B~\cite{qwen2.5}, with LoRA rank, 8) with larger parameter LLMs (Qwen2.5-3B~\cite{qwen2.5}, with higher-rank LoRA, 16) and medical LLMs (BioMistral-7B~\cite{biomistral}) for model tuning. As shown in Fig.~\ref{fig:plug:llm}, larger parameter LLMs yield slight information gains on DEC and READ Pred. However, we observe that the improvements from the medical LLMs were not as pronounced as compared to the other backbone variants—in fact, their performance was at best comparable and at worst.
This limited enhancement may arise from the overlap of knowledge within the medical LLMs and the external RAG knowledge, as both are based on training with open-source data. Additionally, the medical LLMs underwent extra reinforcement learning from human feedback specifically designed for textbook question-answering tasks, which may be less relevant to the healthcare predictions addressed in this study.

\noindent\textbf{Diverse Experiment Settings.} 
To demonstrate robustness across different training settings, we select a training data split of 0.8:0.1:0.1 for testing, following GraphCare~\cite{jianggraphcare} and KARE~\cite{jiangkare}, as shown in Fig.~\ref{fig:train}. Please note that for generative patterns, AUROC and Balance ACC are equivalent, with the baseline derived from the generative pattern directly outputting text instead of logits. Comparisons with competitive baselines indicate that our performance improvements remain robust.
We found that in situations with richer information, such as GraphCare in READ Pred, the discriminant still has an advantage, and does not completely lose its advantage as described in KARE. Our strategy improves accuracy through a greater number of cases, exposing agents to diverse patient patterns, making it more robust than other generative approaches.
\begin{figure}[!h] 
\vspace{-1em}
\centering
\subfigure[DEC Pred]{
\begin{minipage}[t]{0.325\linewidth}
\centering
\includegraphics[width=\linewidth,height=0.75\linewidth]{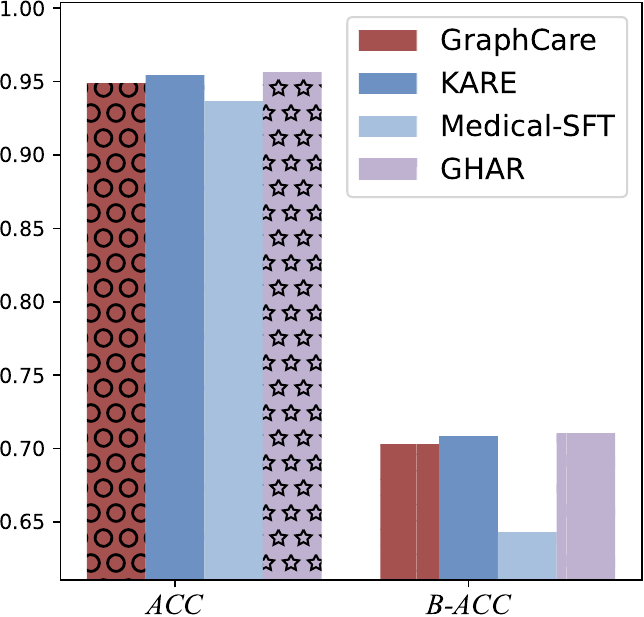}\label{fig:train:dec:bacc}
\end{minipage}%
}%
\subfigure[READ Pred]{
\begin{minipage}[t]{0.325\linewidth}
\centering
\includegraphics[width=\linewidth,height=0.75\linewidth]{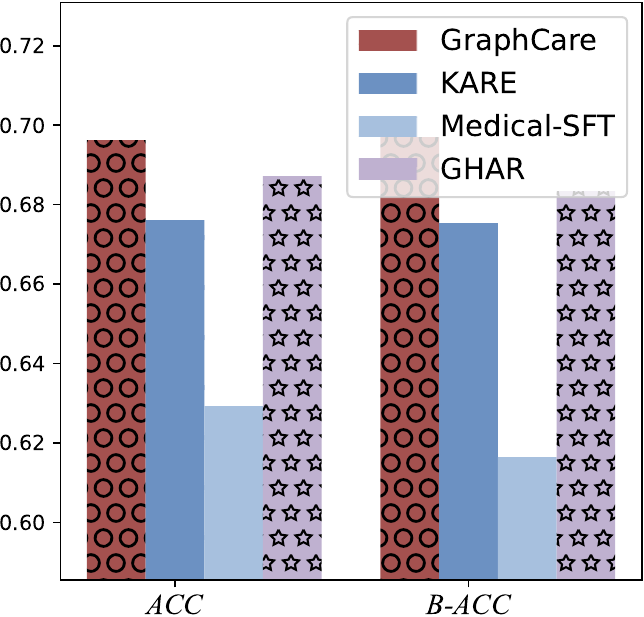}\label{fig:train:rea:bacc}
\end{minipage}%
}%
\subfigure[LOS Pred]{
\begin{minipage}[t]{0.325\linewidth}
\centering
\includegraphics[width=\linewidth,height=0.75\linewidth]{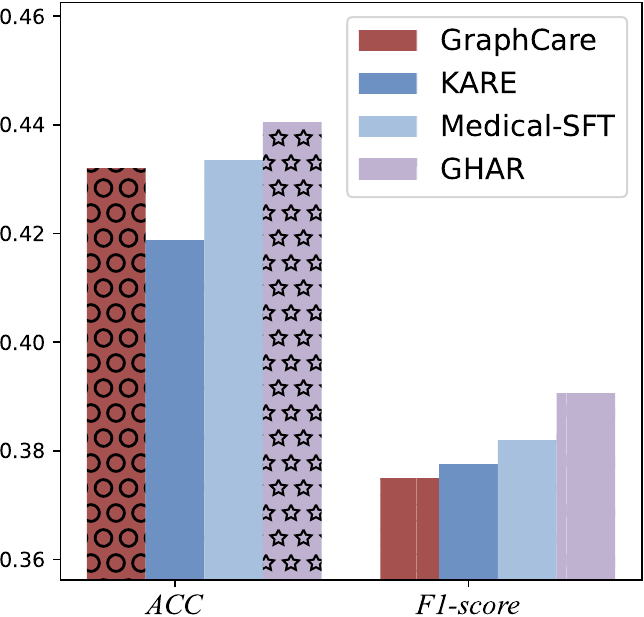}\label{fig:train:los:f1}
\end{minipage}%
}%
\centering
\setlength{\abovecaptionskip}{-0.15cm}   
\setlength{\belowcaptionskip}{-0.1cm}   
\caption{Diverse Training Settings. Please note that in KARE, DEC Pred denotes in-hospital mortality, not within 24h.}
\label{fig:train}
\vspace{-0.3cm}
\end{figure}

\begin{figure}[!h] 
\centering
\subfigure[MIMIC-III-REA$\rightarrow$III-DEC]{
\begin{minipage}[t]{0.45\linewidth}
\centering
\includegraphics[width=\linewidth,height=0.7\linewidth]{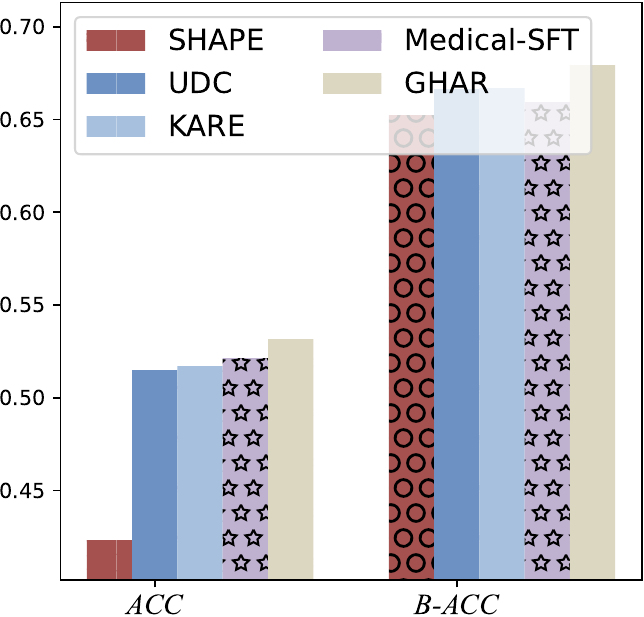}\label{fig:ood:rea-dec}
\end{minipage}%
}%
\subfigure[MIMIC-III-LOS$\rightarrow$IV-LOS]{ 
\begin{minipage}[t]{0.45\linewidth}
\centering
\includegraphics[width=\linewidth,height=0.7\linewidth]{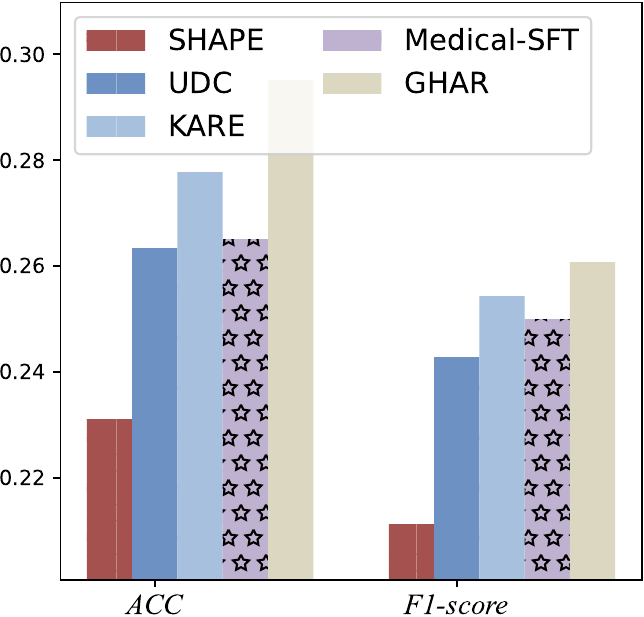}\label{fig:ood:miii-iv} 
\end{minipage}%
}%
\centering
\setlength{\abovecaptionskip}{-0.15cm}   
\setlength{\belowcaptionskip}{-0.1cm}   
\caption{OOD Examination. (a) cross-task scenario. (b) cross-dataset scenario. For both scenarios, we use one domain as the pre-training domain and then directly assess performance on the test set of the other domain.}\label{fig:ood}
\vspace{-0.3cm}
\end{figure}

\subsection{Out-of-Distribution Examination}\label{sec:rob:ood}
We conduct additional out-of-distribution (OOD) tests to verify the zero-shot transferability of our methods. This represents a significant advantage of generative approaches, as traditional discriminative methods adhere to the independent and identically distributed assumption, making it challenging to generalize to scenarios with significant distribution differences~\cite{fan2024survey}. Specifically, we design two types of OOD scenarios: cross-task and cross-dataset. The former involves training on READ and testing on the DEC Pred task, while the latter consists of training on MIMIC-III and testing on MIMIC-IV.
As shown in Fig.~\ref{fig:ood}, both KARE and our method outperform SHAPE by more than 10\%. This improvement is attributed to the generative paradigm's ability to capture the fundamental semantics of entities and labels, going beyond the shallow co-occurrence that discriminative models rely on for prediction, as further validated in Fig.~\ref{fig:case:case:exp}.
Moreover, in both tasks, our approach maintains an advantage over KARE and Medical-SFT, which can be attributed to the collaborative interplay between the two agents. This collaboration helps bridge the gap between retrieval and generation, allowing for an iterative selection process that accommodates a greater volume of external information. Meanwhile, RAG effectively addresses knowledge gaps in OOD scenarios by dynamically retrieving relevant and essential information from external knowledge bases. In contrast, Medical-SFT relies on fixed training data, limiting its generalization capability. As a result, GHAR demonstrates superior performance in OOD situations.
\begin{figure}[!h] 
\centering
\subfigure[DEC Pred (B-ACC)]{
\begin{minipage}[t]{0.325\linewidth}
\centering
\includegraphics[width=\linewidth,height=0.75\linewidth]{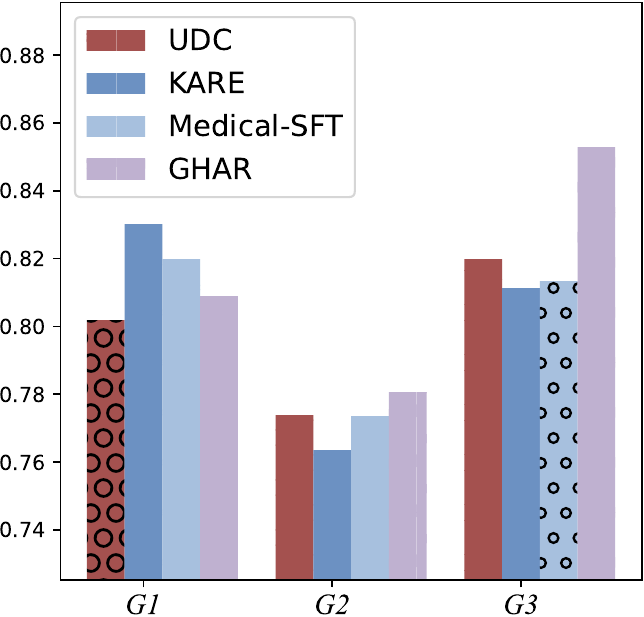}\label{fig:gr:dec:bacc}
\end{minipage}%
}%
\subfigure[READ Pred (B-ACC)]{
\begin{minipage}[t]{0.325\linewidth}
\centering
\includegraphics[width=\linewidth,height=0.75\linewidth]{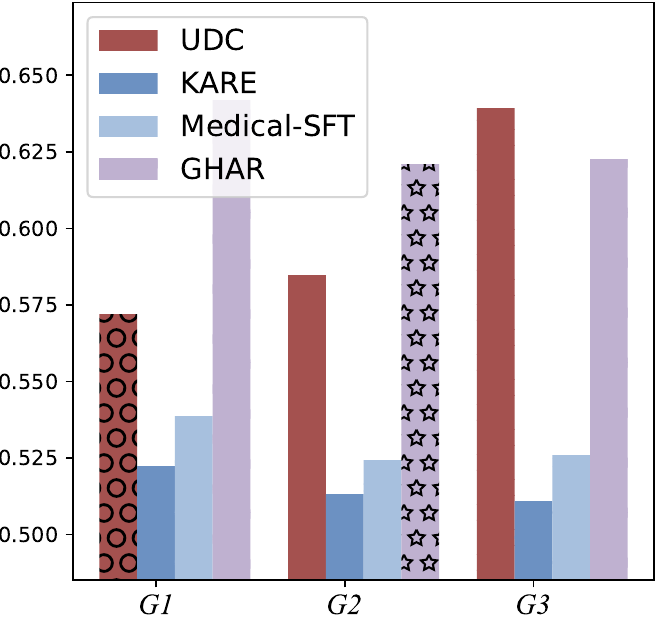}\label{fig:gr:rea:bacc}
\end{minipage}%
}%
\subfigure[LOS Pred (F1-score)]{
\begin{minipage}[t]{0.325\linewidth}
\centering
\includegraphics[width=\linewidth,height=0.75\linewidth]{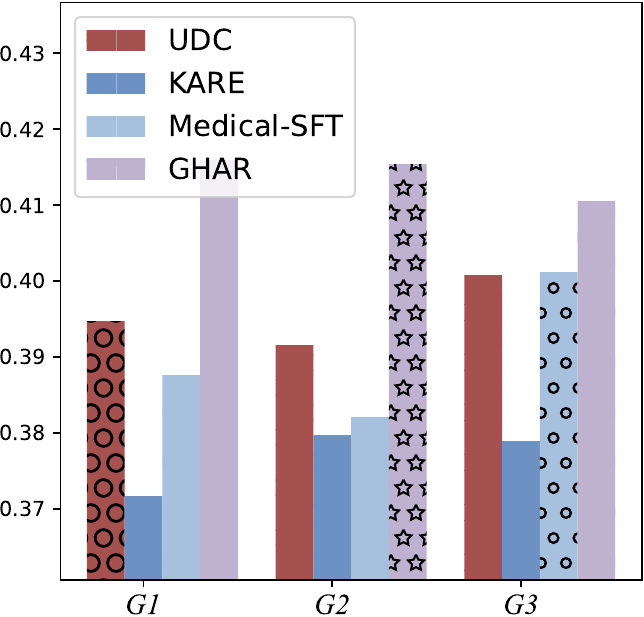}\label{fig:gr:los:f1}
\end{minipage}%
}%
\centering
\setlength{\abovecaptionskip}{-0.15cm}   
\setlength{\belowcaptionskip}{-0.1cm}   
\caption{Group-wise Analysis. G1-G3 refer to groups based on disease rarity:G1: [0, 1/3),
G2: [1/3, 2/3),
G3: [2/3, 1].}
\label{fig:gr}
\vspace{-0.5cm}
\end{figure} 
\subsection{Group-Wise Performance}\label{sec:rob:gr}
In real medical scenarios, the types of diseases among patients are often imbalanced, with common illnesses dominating the dataset~\cite{zhao2025}.
Following~\cite{liu2025generalist,zhao2025}, we categorize patients based on the rarity of their diseases into groups G1-G3, with G1 representing the rarest group. As shown in Fig.~\ref{fig:gr}, the overall performance of both discriminative and generative methods improves with the prevalence of diseases (G2$\rightarrow$G3). For the discriminative methods, this improvement is likely due to the strong co-occurrence present in the group. In contrast, generative methods may benefit from the prevalence of common diseases during the SFT / RL learning phase.
Notably, the distribution of generative methods is more uniform overall, especially in our approach. For instance, on the READ prediction task, our algorithm demonstrates superior performance on group G1 without being adversely affected by the more frequently occurring group G3. This uniformity can be attributed to two factors: first, it captures the semantic similarity between entities regardless of co-occurrence.  This is crucial, as the semantics derived from the text are independent of the dataset's interactions, thus mitigating the effects of skewed distributions.
Second, our model explores a wider range of potentially relevant entities during the deep thinking process, thus expanding the boundaries of knowledge. This capability enhances the effectiveness of our method across various disease categories. To sum up, adopting a fairer approach for the rare diseases group can lead to improved clinical outcomes by ensuring that patients receive more accurate diagnoses and tailored treatment options~\cite{zhao2025}.

\subsection{Semantic Understanding}\label{app:sum}
In addition to evaluating healthcare generation, we conduct a thorough assessment of free-text generation performance, specifically focusing on question-answering (QA) tasks. To achieve this, we leverage the MMLU-Clinical~\cite{kim2025questioning} and MIMIC-IV-Ext-BHC datasets~\cite{aali2024mimic}, where the input comprises rigorous clinical questions or original clinical notes. The objective is to generate multiple-choice answers or concise summaries formulated by expert clinicians. This task serves as a robust measure of the model's ability to produce coherent outputs and its depth of understanding in complex medical contexts.
As demonstrated in Table~\ref{tab:sum}, our performance metrics in the medical QA tasks, including accuracy (ACC) and F1-score, indicate optimal results. Moreover, in the summary task, our ROUGE and SARI scores~\cite{ermakova2019survey,zhumed} for this demanding free-text generation task are significantly elevated. In contrast, the MedRAG and KARE frameworks, which utilize the single-round retrieval approach, encounter substantial challenges in adapting to the deeper knowledge requirements inherent in these complex semantic understanding scenarios. This further highlights the superiority of our agentic RAG strategy, which proactively engages with semantic intricacies to ensure precise comprehension of questions and optimal external knowledge integration.
\begin{table}[!ht]
\centering
\setlength{\abovecaptionskip}{-0.05cm}   
\setlength{\belowcaptionskip}{-0.1cm}   
\caption{Semantic Understanding. Compared to traditional Healthcare Prediction tasks, this approach no longer focuses on understanding patient states; instead, it emphasizes understanding the diverse medical problems and the more challenging task of free-text generation.}\label{tab:sum}
\resizebox{0.49\textwidth}{!}{
\begin{tabular}{c|cc||ccc} 
\toprule
\multirow{2}{*}{Methods} & \multicolumn{2}{c||}{QA} & \multicolumn{3}{c}{Summary}                          \\ 
\cline{2-6}
                         & ACC$\uparrow$    & F1-score$\uparrow$        & ROUGE-1$\uparrow$ & ROUGE-L$\uparrow$ & SARI$\uparrow$  \\ 
\hline
DeepSeekR1-7B            & 0.3952 & 0.3921          & 0.2193         & 0.1188            & 0.3839          \\
Biomistral-7B              & 0.2610 & 0.3572          & 0.1720         & 0.1016            & 0.3841          \\
LightRAG                 & 0.6913 & 0.6915          & 0.2719         & 0.1557            & 0.3995          \\
Search-R1                & 0.7031 & 0.7039          & 0.2896         & 0.1611            & 0.3937          \\
MedRAG        & 0.6991 & 0.7003          & 0.2642         & 0.1500            & 0.3990 \\
KARE                & 0.7011 & 0.7015          & 0.2966         & 0.1758            & 0.4228          \\
Medical-SFT                   & 0.5992 & 0.6058          & 0.2458         & 0.1399            & 0.3859          \\ 
\hline
\textbf{Ours}                     & \textbf{0.7167} & \textbf{0.7182}          & \textbf{0.3090}         & \textbf{0.1808}            & \textbf{0.4307}          \\
\bottomrule
\end{tabular}}
\vspace{-1em}
\end{table}

\subsection{Complexity and Distributed Optimization}\label{sec:rob:com} 
As shown in Fig.~\ref{fig:time:com}, we want to emphasize that under the same generative RAG paradigm, our method is more efficient. Operating directly on raw documents rather than on refined KG entities and relations, Search-R1 requires a more extensive exploration process and tokens, making it more susceptible to the ``loss-in-the-middle" phenomenon.
MedRAG and KARE perform retrieval indiscriminately across all queries, whereas our Agent-Top adaptively determines whether to invoke RAG based on specific needs, significantly reducing the costs associated with vector matching.
Unlike KARE, we do not require synchronizing the output of the CoT at the training stage, which decreases the number of decoding operations needed—a process that can be very time-consuming when performed serially. 
Additionally, our two agents can essentially utilize the same LLM for optimization without introducing extra computational burden, maintaining a complexity level consistent with that of others during the inference phase.

Moreover, in our scenario, the input of a patient's complete longitudinal record within the same batch can be highly imbalanced, and the samples requiring the retrieval process are time-consuming. Therefore, performing LLM generation and retrieval sequentially for a batch is not advisable, as it results in slower speeds.
This issue is not exclusive to our method but is common across most RAG-based approaches. To mitigate this, we optimize the overall computational framework through distributed strategies.
We employ Ray for distributed request handling and utilize gunicorn to implement an asynchronous request system, as depicted in Fig.~\ref{fig:time:fra}.  Meanwhile, our retrieval process operates only on selected meta-path partitions, thereby reducing both the scope of indexing and the computational time required. As demonstrated in Fig.~\ref{fig:time:dis}, these three strategies significantly reduce inference time, thereby enhancing the industrial applicability of our approach.

\begin{figure}[!h] 
\centering
\subfigure[Complexity]{
\begin{minipage}[t]{0.325\linewidth}
\centering
\includegraphics[width=\linewidth,height=0.9\linewidth]{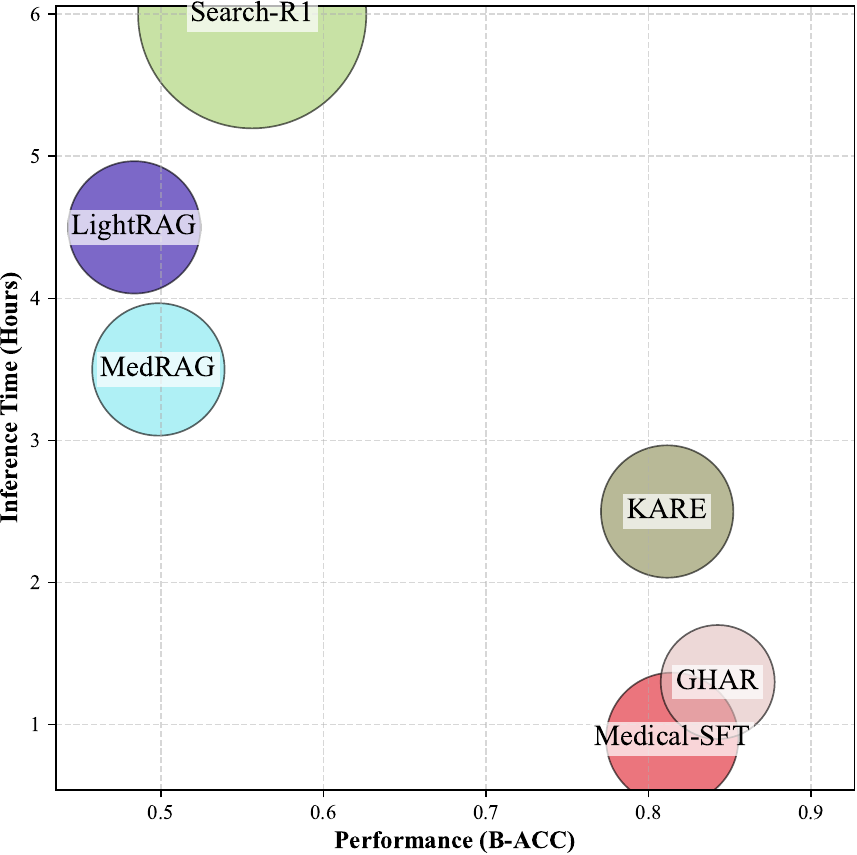}\label{fig:time:com}
\end{minipage}%
}%
\subfigure[Distributed Design]{
\begin{minipage}[t]{0.325\linewidth}
\centering
\includegraphics[width=\linewidth,height=0.9\linewidth]{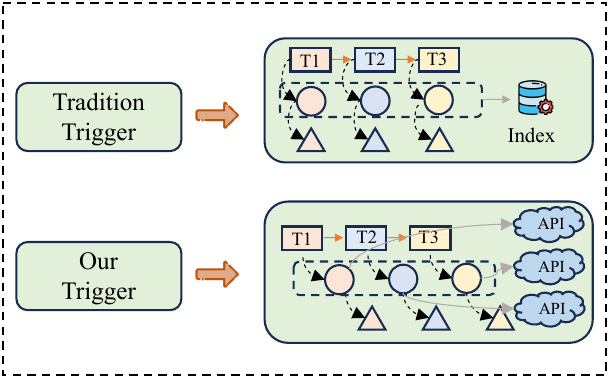}\label{fig:time:fra}
\end{minipage}%
}%
\subfigure[Time Reduction]{
\begin{minipage}[t]{0.325\linewidth}
\centering
\includegraphics[width=\linewidth,height=0.9\linewidth]{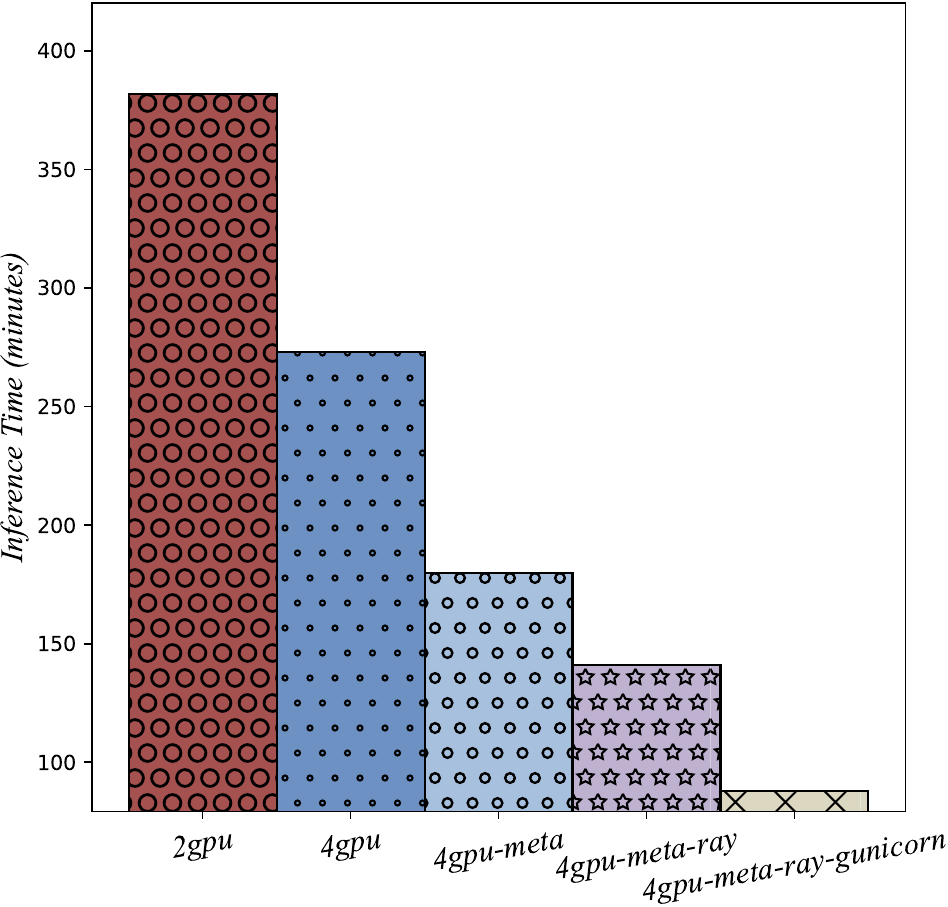}\label{fig:time:dis}
\end{minipage}%
}%
\centering
\setlength{\abovecaptionskip}{-0.15cm}   
\setlength{\belowcaptionskip}{-0.1cm}   
\caption{Time Complexity. To demonstrate practicality and fairness, for Fig.~\ref{fig:time:com} and~\ref{fig:time:dis}, we test the inference time for 1 epoch (MIMIC-III, DEC Pred) on a machine equipped with four RTX 3090 GPUs.}\label{fig:time}
\vspace{-0.3cm}
\end{figure}

\subsection{Case Studies}\label{sec:rob:case}
We conduct case studies to illustrate our rationale.

\noindent\textbf{Iterative Deep Retrieval.} 
We present the additional thinking processes of competitive RAG baselines and our algorithm in Fig.~\ref{fig:case:ret:num}, which include the invocation of RAG operations and LLM-based reasoning. The results clearly demonstrate that our algorithm achieves optimal performance with the minimum number of additional thinking processes.
To investigate the advantages of adaptive iteration, we replace the original adaptive decision-making mechanism of Agent-Top with a strategy that employs a fixed number of RAG calls. The results in Fig.~\ref{fig:case:ret:fix} show a decline in performance, underscoring the importance of adaptively determining the depth of reasoning. A fixed number of iterations may lead to either overthinking or underthinking, consequently reducing accuracy for some questions.
Additionally, our findings indicate that GHAR effectively identifies retrieval needs, mitigating potential issues of knowledge gaps or redundancy. This is further confirmed in Fig.~\ref{fig:case:ret:prune}, where eliminating unnecessary RAG pipelines enhances the baselines' performance on healthcare predictions.
\begin{figure}[!h] 
\vspace{-1cm}
\centering
\subfigure[Extra Thinking]{
\begin{minipage}[t]{0.325\linewidth}
\centering
\includegraphics[width=\linewidth,height=0.9\linewidth]{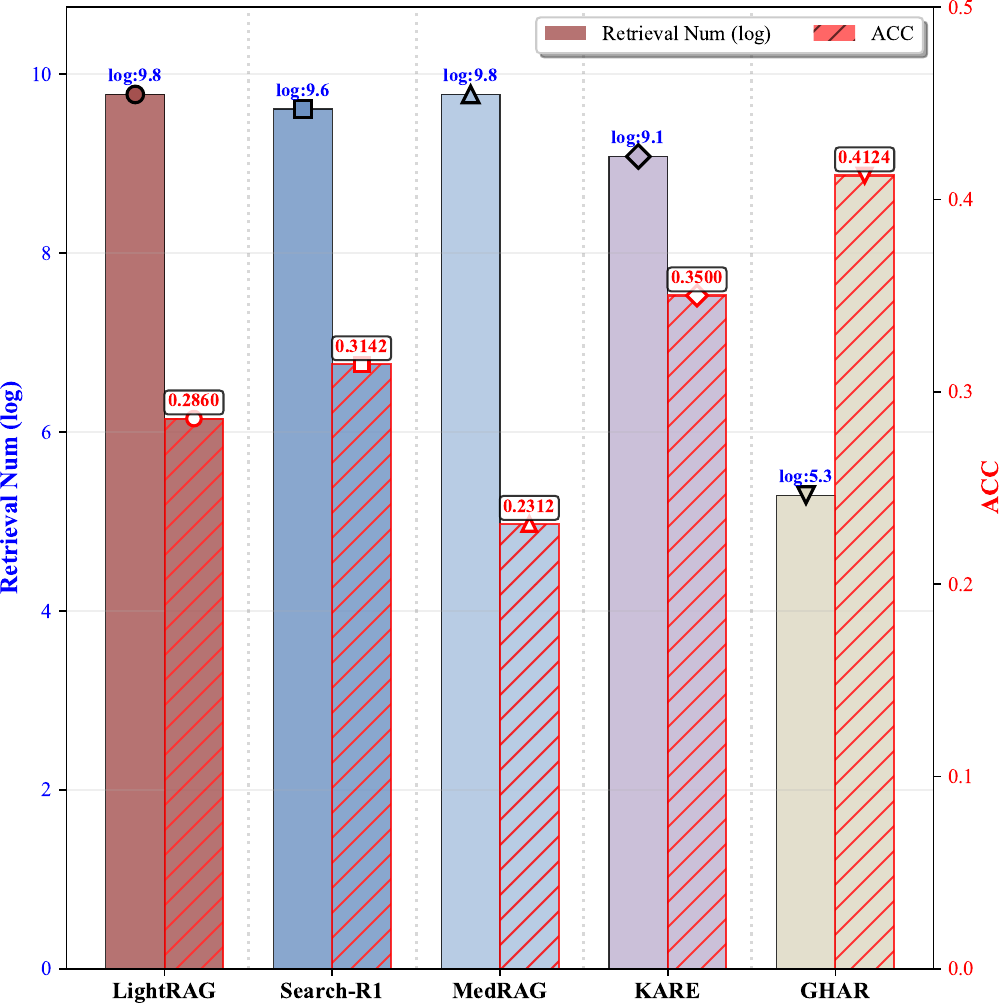}\label{fig:case:ret:num}
\end{minipage}%
}%
\subfigure[Increased Iterations]{
\begin{minipage}[t]{0.325\linewidth}
\centering
\includegraphics[width=\linewidth,height=0.9\linewidth]{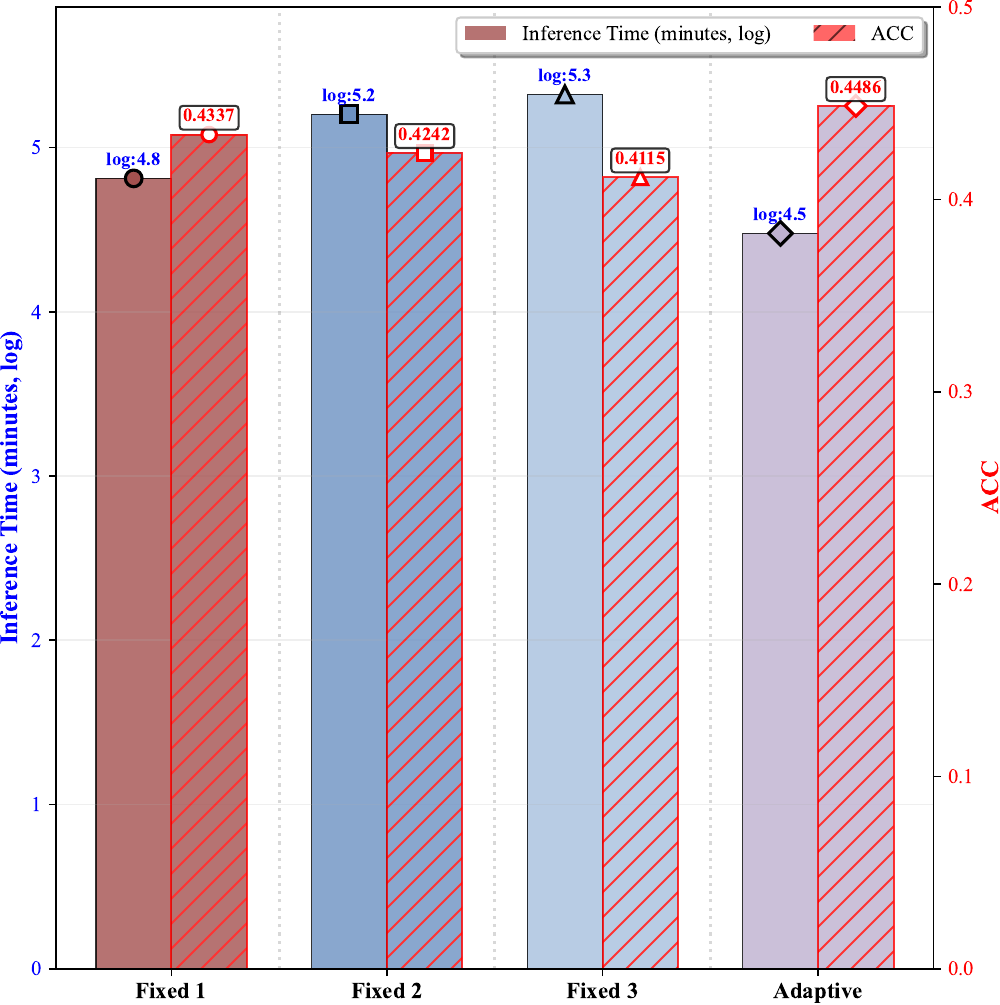}\label{fig:case:ret:fix}
\end{minipage}%
}%
\subfigure[RAG Pruning]{
\begin{minipage}[t]{0.325\linewidth}
\centering
\includegraphics[width=\linewidth,height=0.9\linewidth]{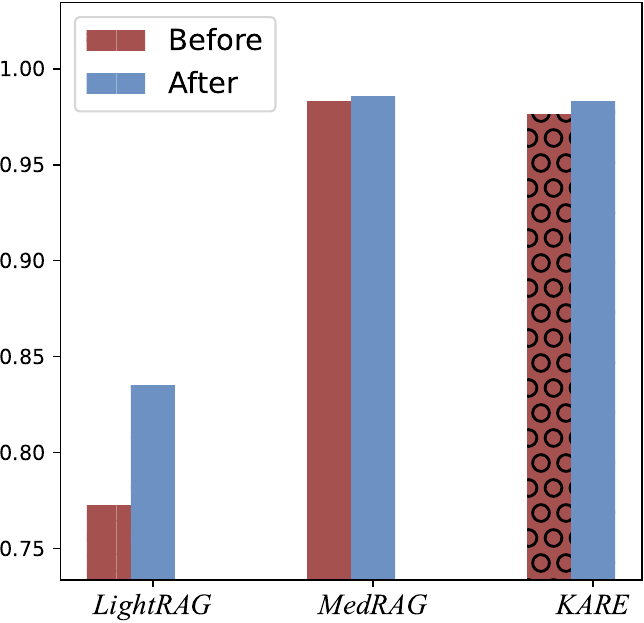}\label{fig:case:ret:prune}
\end{minipage}%
}%
\centering
\setlength{\abovecaptionskip}{-0.15cm}   
\setlength{\belowcaptionskip}{-0.1cm}   
\caption{Iterative Deep Retrieval. (a)  Extra Thinking \& Performance on LOS Pred (MIMIC-III). (b) Examine the performance variation brought by growth iterations. (c) During the SFT phase of these algorithms, we remove the cases identified by our method as not requiring the RAG process and replace them with corresponding query-answer pairs as training data.}
\label{fig:case:ret}
\vspace{-0.1cm}
\end{figure}


\noindent\textbf{Illustrative Cases.}
In Fig.~\ref{fig:case:case:kg}, we compare the knowledge extraction approaches of different algorithms. Our algorithm can correctly answer questions without the need for explicit knowledge extraction. In contrast, both Search-R1 and KARE encounter coarse-grained knowledge, which may lead the model to fall into a loss-in-middle scenario, resulting in incorrect answers.
Another case study in Fig.~\ref{fig:case:case:dec} illustrates our algorithm's RAG-based extraction process. Our method conducts a two-iteration process in which the RAG component accesses only the partitions corresponding to the meta-path ``(`drug', `drug\_protein', `gene/protein’)", significantly reducing indexing time. Furthermore, from the perspective of sub-query reasoning, our algorithm iteratively identifies and addresses existing knowledge gaps (i.e., the second subquery), thereby enhancing overall performance.

Additionally, we emphasize that generative algorithms offer greater explainability compared to traditional discriminative algorithms. They not only provide rationales but also effectively capture the underlying correlations among different labels. This is convincingly demonstrated in Fig.~\ref{fig:case:case:exp}, which illustrates the representational similarities of responses across various labels. Specifically, the representation distance between responses for adjacent days is smaller, indicating a higher degree of similarity. This approach is cognitively aligned, as it captures the semantic relationships between labels. In contrast, discriminative methods treat labels as discrete, independent entities, merely predicting the most probable one without comprehending their underlying similarities.

\begin{figure}[!h] 
\vspace{-1em}
\centering
\subfigure[Inference Comparison]{
\begin{minipage}[t]{0.45\linewidth}
\centering
\includegraphics[width=\linewidth,height=0.85\linewidth]{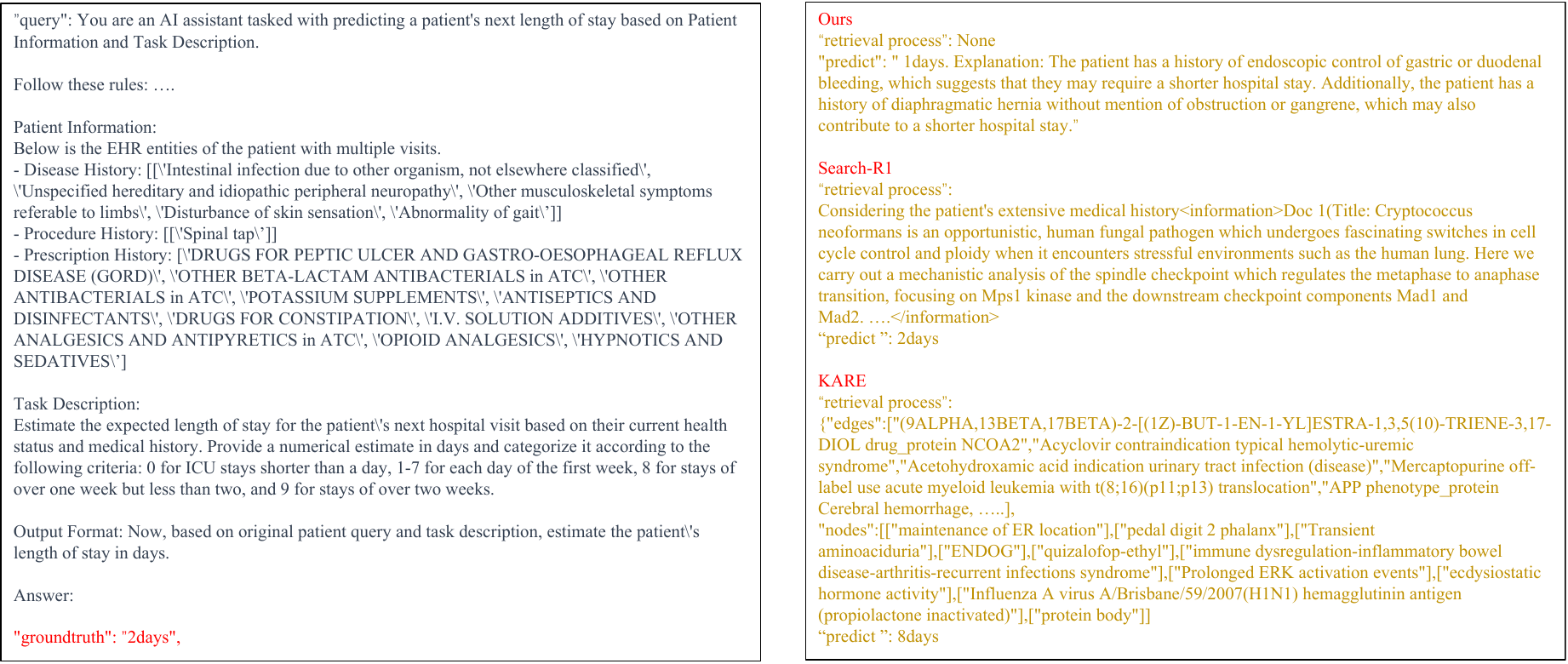}\label{fig:case:case:kg}
\end{minipage}
}%
\subfigure[Our RAG Pipeline]{
\begin{minipage}[t]{0.45\linewidth}
\centering
\includegraphics[width=\linewidth,height=0.85\linewidth]{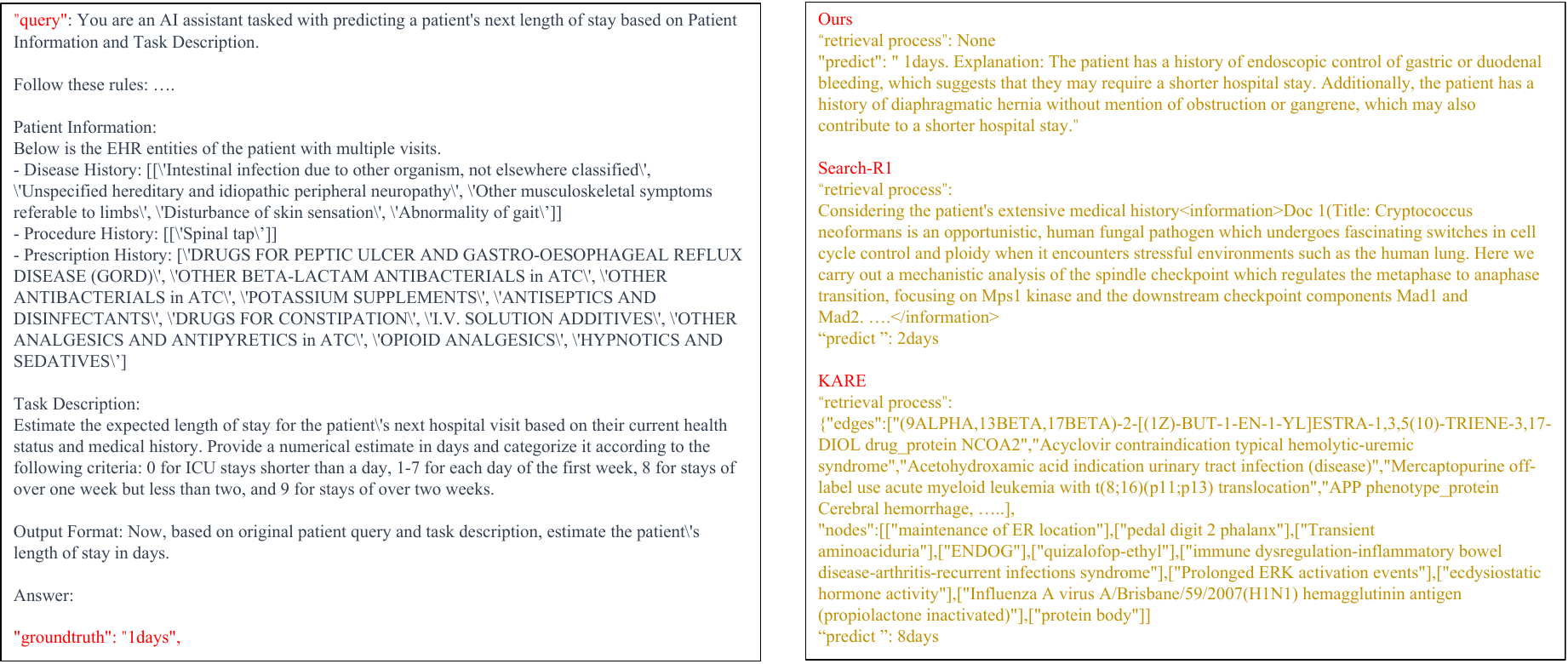}\label{fig:case:case:dec}
\end{minipage}%
}%
\vfill
\subfigure[Explainable Response]{
\begin{minipage}[t]{0.75\linewidth}
\centering
\includegraphics[width=\linewidth,height=0.45\linewidth]{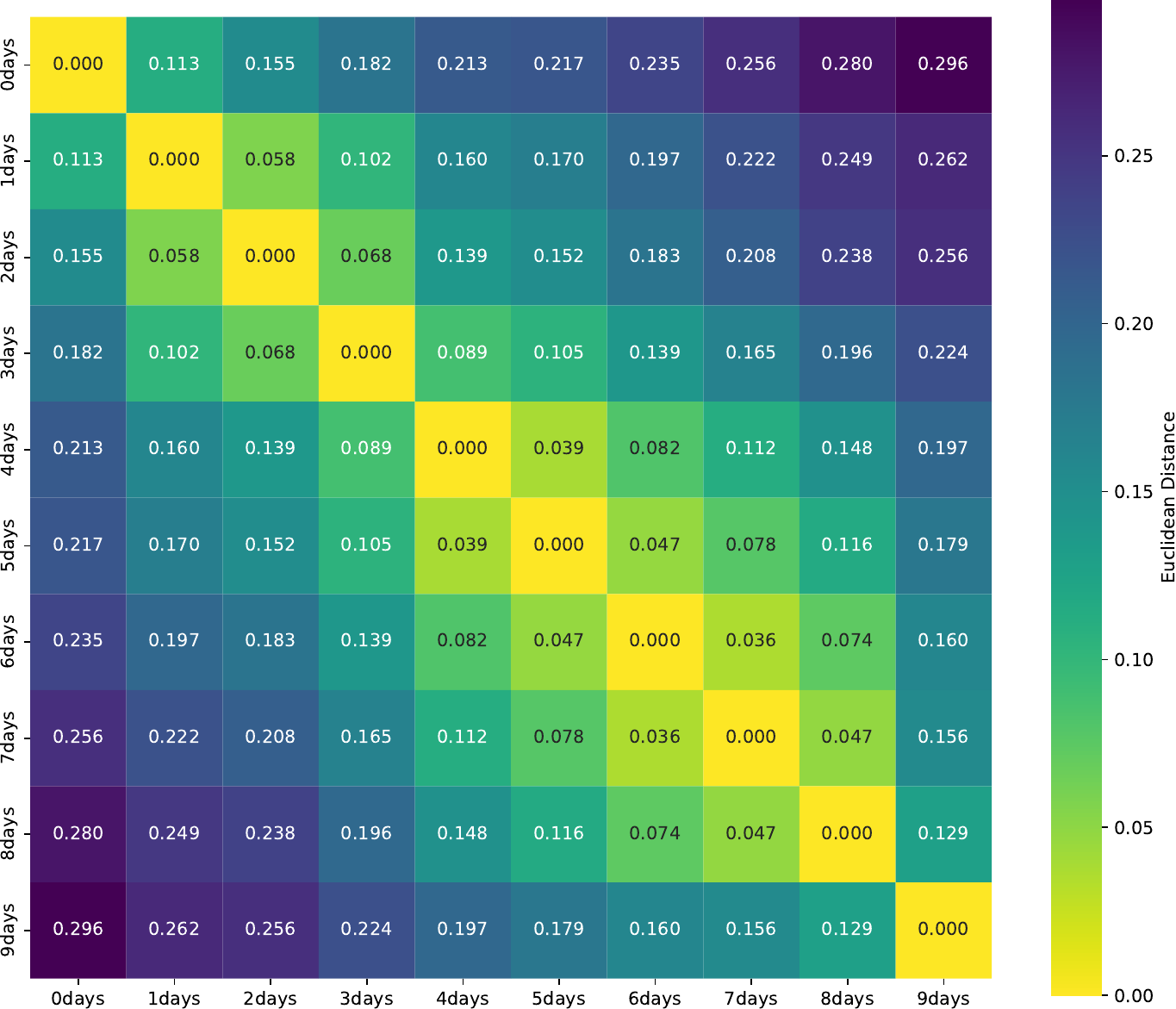}\label{fig:case:case:exp}
\end{minipage}%
}%
\centering
\setlength{\abovecaptionskip}{-0.15cm}   
\setlength{\belowcaptionskip}{-0.1cm}   
\caption{Illustrative Cases. (a) Inference process for final prediction. (b) Another case of our retrieval process. (c) Explainable label relationships (MIMIC-III, LOS Pred).} 
\label{fig:case:case}
\vspace{-0.3cm}
\end{figure}

\subsection{Hyper-parameter Tests}\label{sec:rob:hyper}
We examine the essential hyperparameters in our experiments to achieve optimal performance. 
\begin{figure}[!h] 
\centering
\subfigure[Meta-path Size]{
\begin{minipage}[t]{0.325\linewidth}
\centering
\includegraphics[width=\linewidth,height=0.85\linewidth]{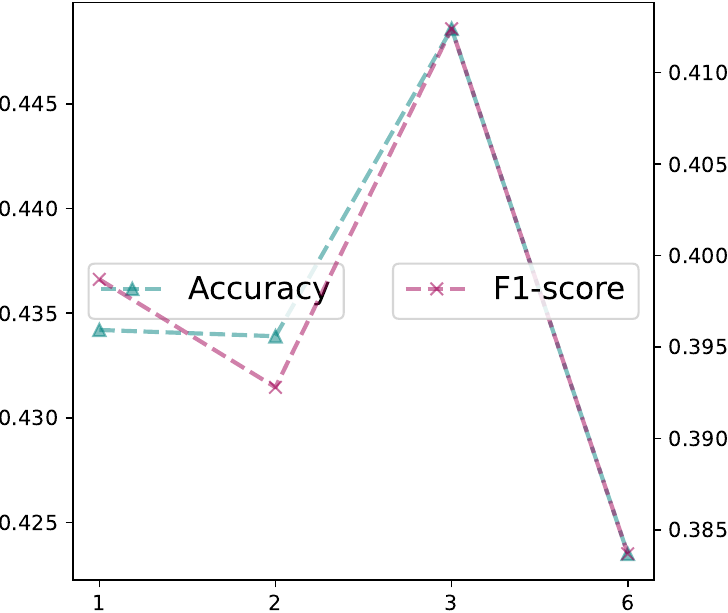}
\label{fig:hyper:meta}
\end{minipage}%
}%
\subfigure[Top-K Recall]{
\begin{minipage}[t]{0.325\linewidth}
\centering
\includegraphics[width=\linewidth,height=0.85\linewidth]{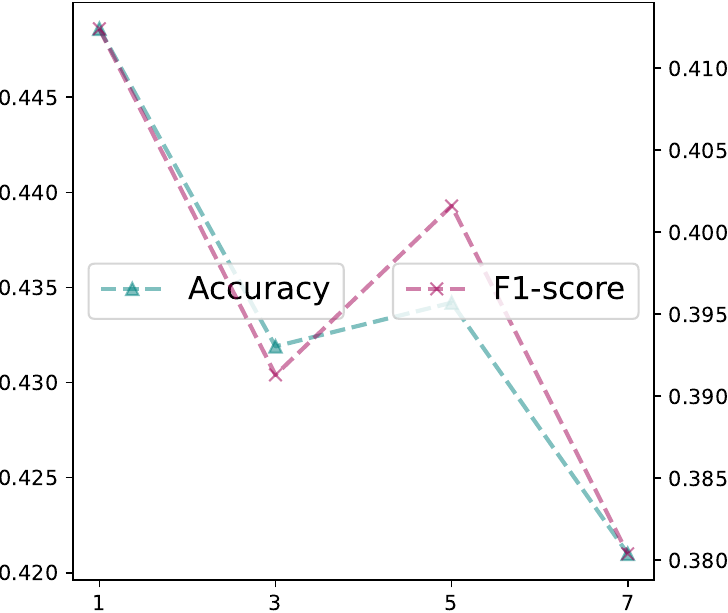}
\label{fig:hyper:topk}
\end{minipage}%
}%
\subfigure[ORM Weight]{
\begin{minipage}[t]{0.325\linewidth}
\centering
\includegraphics[width=\linewidth,height=0.85\linewidth]{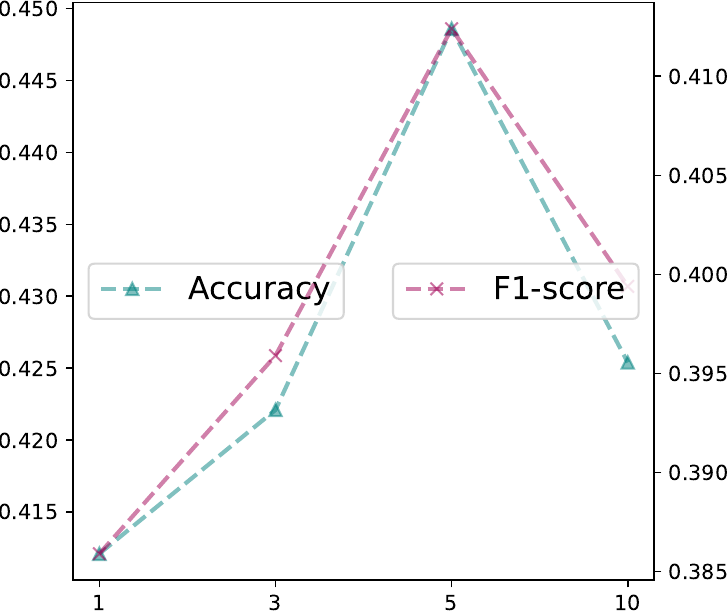}
\label{fig:hyper:orm}
\end{minipage}%
}%
\centering
\setlength{\abovecaptionskip}{-0.15cm}   
\setlength{\belowcaptionskip}{-0.1cm}   
\caption{Hyper-parameter Tests. Here, we take the LOS Pred task (MIMIC-III) as an example.}
\label{fig:hyper}
\vspace{-0.1cm}
\end{figure}

\noindent\textbf{Maximum Number of Meta-paths $|\mathcal{\tilde{O}}_{\text{c}}|$.}
It represents the breadth of meta-path selection. For complex issues, insufficient breadth may hinder thorough problem decomposition, potentially leading to unreasonable results. Conversely, excessive selection can introduce redundancy or irrelevant knowledge. Our experiments in Fig.~\ref{fig:hyper:meta} indicate that optimal performance is achieved when $|\mathcal{\tilde{O}}_{\text{c}}|=3$. This suggests a balanced approach to the quantity of meta-paths, enabling effective problem-solving without unnecessary complexity.

\noindent\textbf{Top-K Recall $N$.} It indicates the number of nodes and edges matched during the RAG, which is crucial in GraphRAG-related work. A higher quantity provides a broader knowledge base, offering sufficient resources for Agent-Low to generate summaries. However, excessively high values may introduce irrelevant nodes and relationships from the KG, effectively adding noise and impacting subsequent knowledge refinement.
According to Fig.~\ref{fig:hyper:topk}, performance improves when $N=1$.

\noindent\textbf{ORM Reward Weight $\eta$.} The ORM reward is a result-based incentive that serves as a shared objective for both models, in contrast to a standalone cost. A larger reward encourages collaboration between the two agents, while a smaller reward may lead the models to focus on their individual roles.
As shown in Fig.~\ref{fig:hyper:orm}, optimal performance is achieved when $\eta=5$. Larger values may cause the sub-agents to neglect their own penalties, negatively impacting the stability of the reinforcement learning process as a whole.

\section{Conclusion}\label{sec:rob:con}
In this paper, we introduce GHAR, a framework that leverages hierarchical agentic RAG to enhance the performance of generative healthcare prediction. Our innovative design incorporates two distinct agents, Agent-Top and Agent-Low, which collaboratively determine when to engage in RAG and what information to retrieve. This iterative retrieval process explicitly integrates a reasoning chain into the model's decision-making framework, fostering more informed predictions.
We also apply principles of multi-agent reinforcement learning, reformulating our approach as a MDP. Through a diverse reward structure, we achieve unified optimization of both the retrieval and generation modules, ensuring their seamless collaboration. Extensive experiments and robust analyses demonstrate the strong performance and interpretability of our model across various scenarios.
Despite these advancements, our model has certain limitations. A key area for future work lies in designing more intuitive process rewards that could further enhance model performance.

\section{Prompt Templates}\label{app:prompt}
We provide the corresponding prompt templates.
\begin{figure}[!ht] 
\centering
\includegraphics[width=0.85\linewidth]{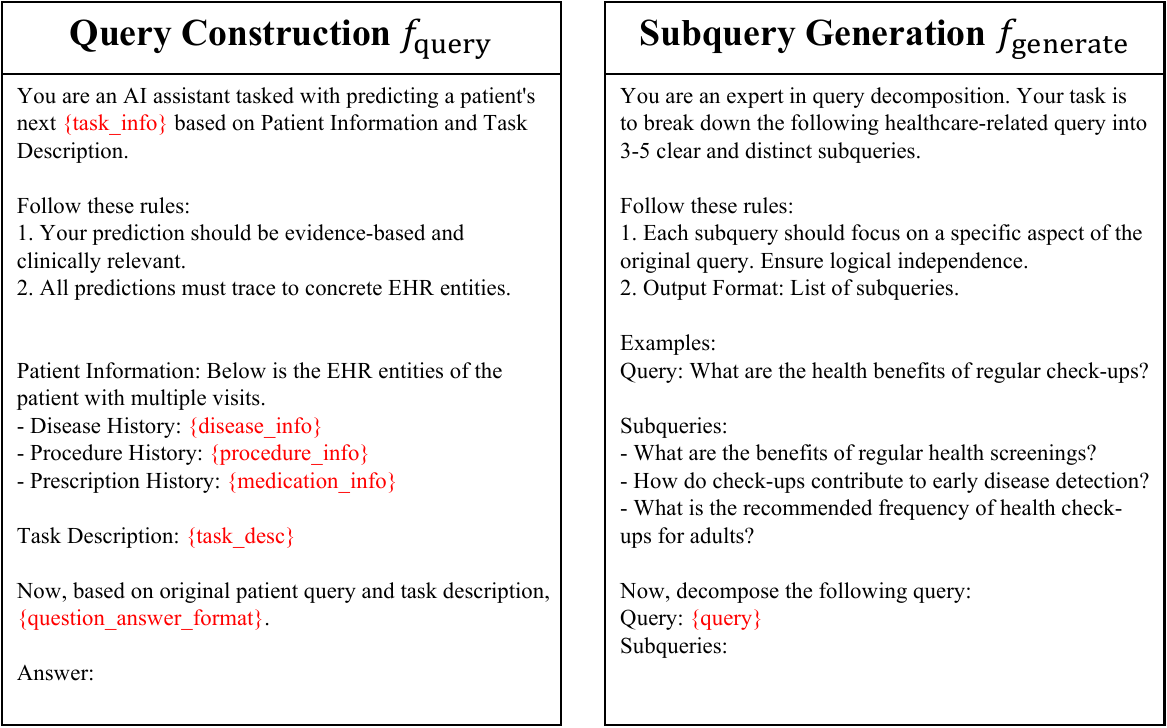}
\caption{Prompt template for section~\ref{sec:med:prompt}. The areas highlighted in red indicate the variables that are passed in. More task information and answer format can be found in Fig.~\ref{fig:temp5}.}
\label{fig:temp1}
\end{figure} 
\begin{figure}[!ht] 
\centering
\includegraphics[width=0.85\linewidth]{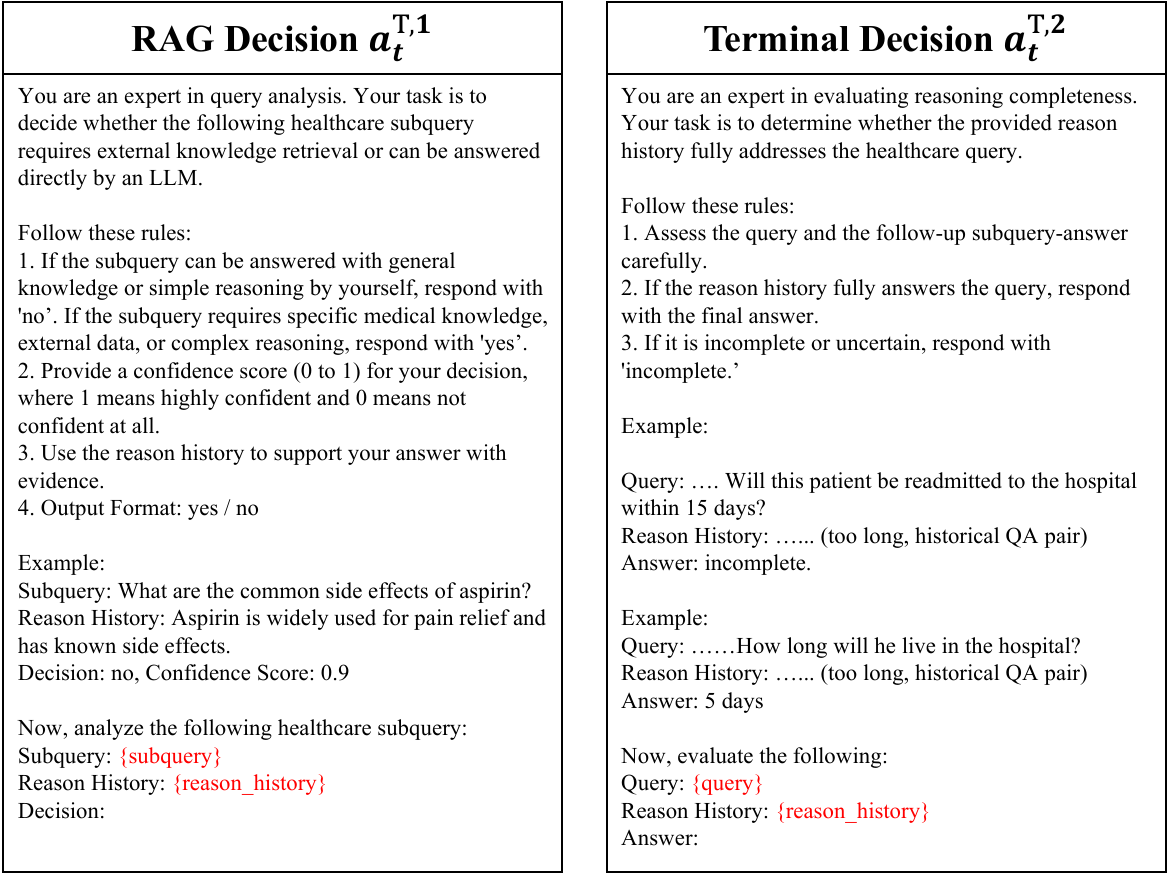}
\caption{Prompt template for Agent-Top decision in section~\ref{sec:med:atop}.}
\label{fig:temp2}
\end{figure} 
\begin{figure}[!ht] 
\centering
\includegraphics[width=0.85\linewidth]{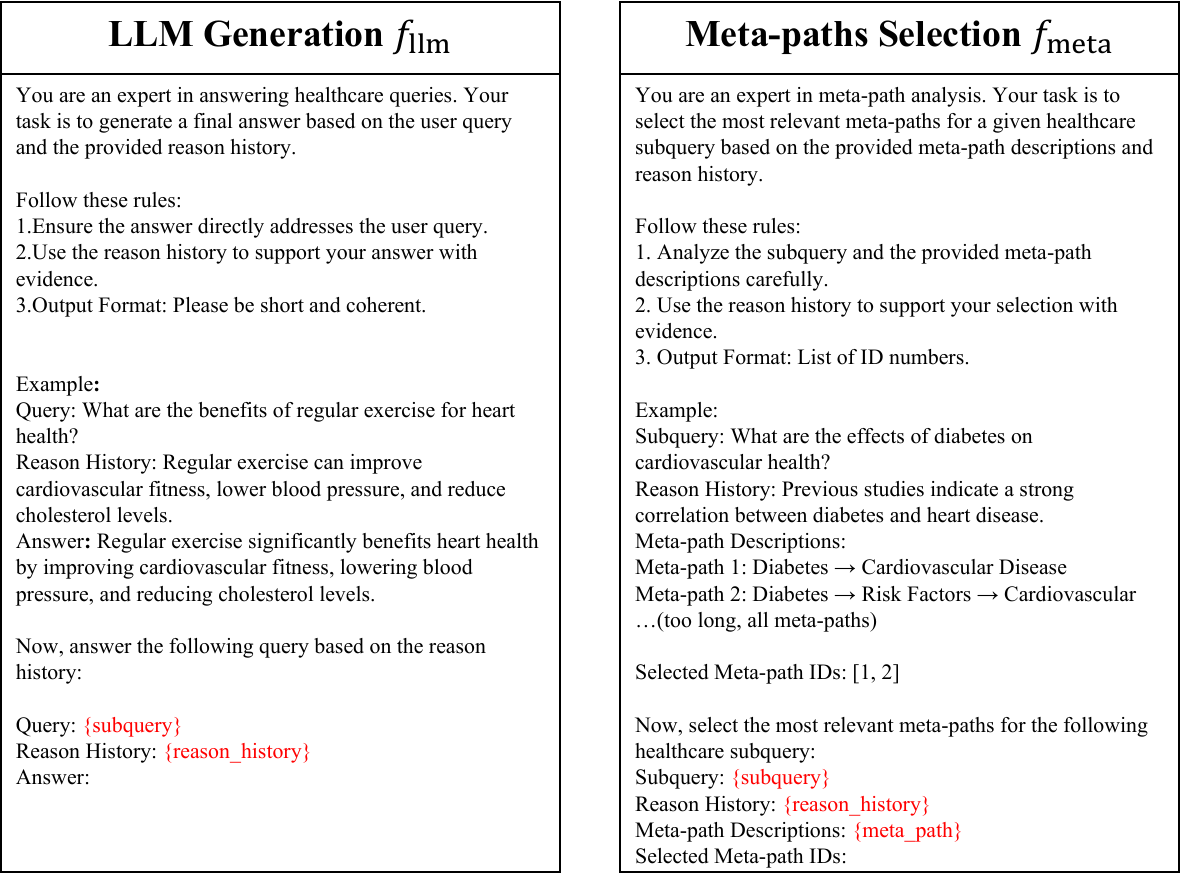}
\caption{Fine-grained prompt template for Agent-Top action in section~\ref{sec:med:atop}.}
\label{fig:temp3}
\end{figure} 
\begin{figure}[!ht] 
\centering
\includegraphics[width=0.85\linewidth]{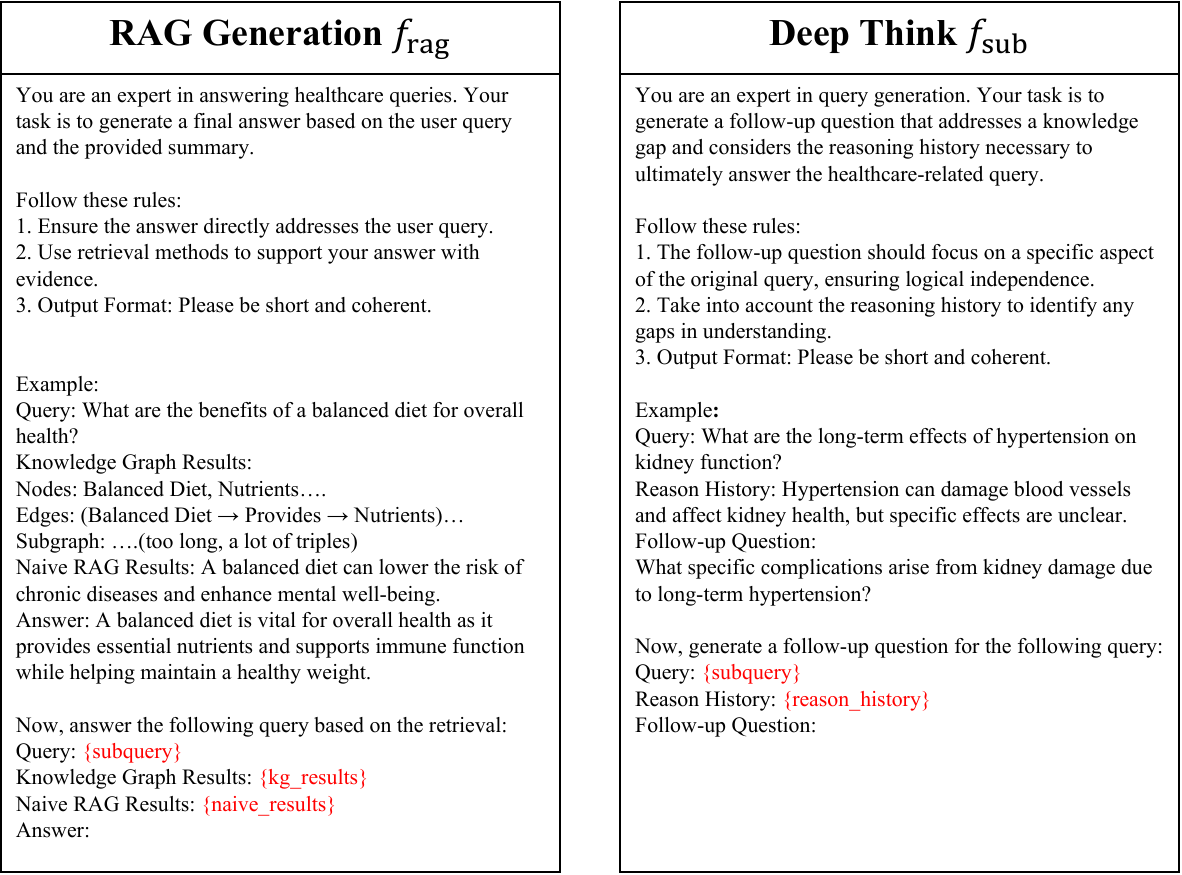}
\caption{Prompt template for Agent-Low action in section~\ref{sec:med:alow} and deep thinking in section~\ref{sec:med:pre}.}
\label{fig:temp4}
\vspace{-0.3cm}
\end{figure} 
\begin{figure}[!ht] 
\centering
\includegraphics[width=0.85\linewidth]{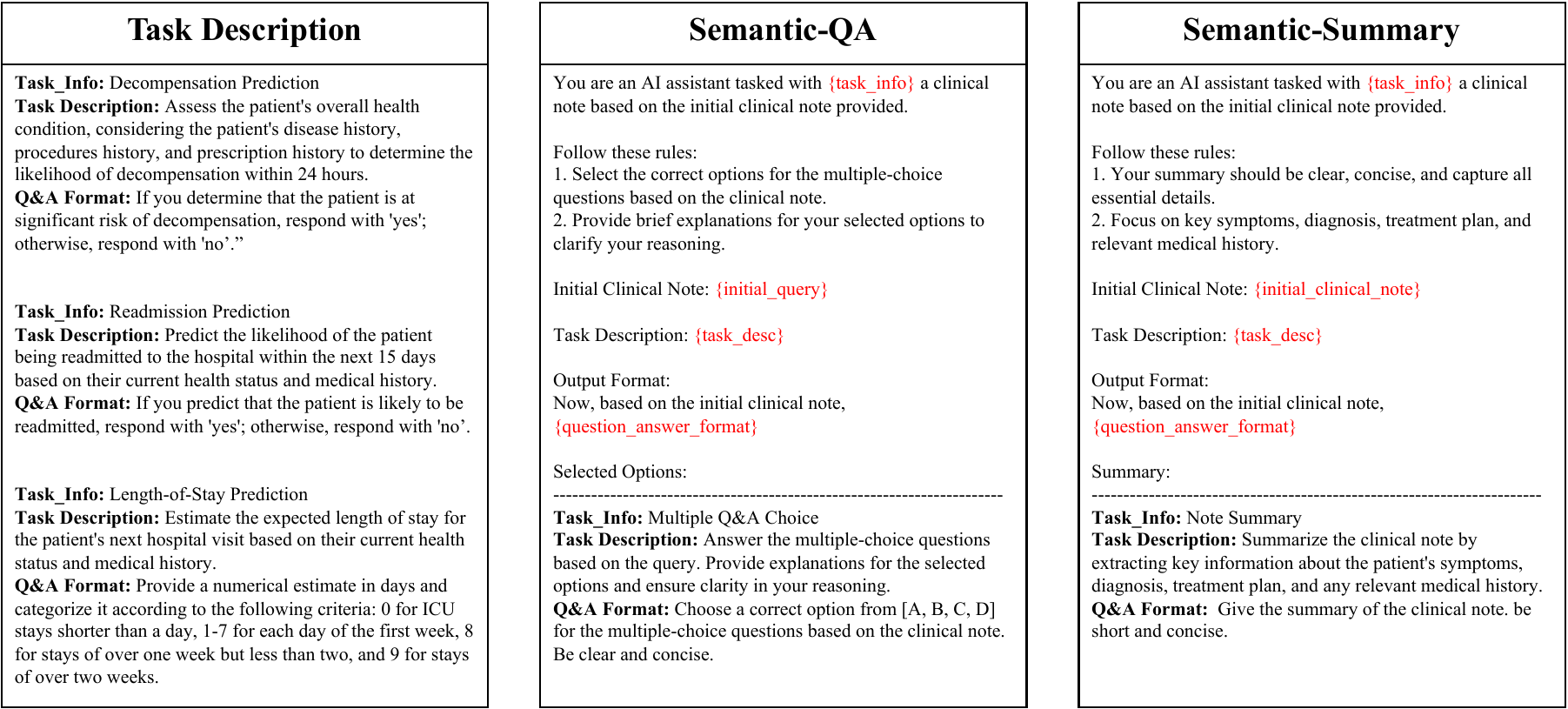}
\caption{Prompt template for task description \& semantic understanding in section~\ref{sec:rob}.}
\label{fig:temp5}
\end{figure} 

\newpage

\bibliographystyle{IEEEtran}

\bibliography{main}

\end{document}